\documentclass{article} 
\usepackage{iclr2023_conference,times}

\usepackage{amsmath,amsfonts,bm}

\def\eqref#1{equation~\ref{#1}}

\def\1{\bm{1}}

\DeclareMathAlphabet{\mathsfit}{\encodingdefault}{\sfdefault}{m}{sl}
\SetMathAlphabet{\mathsfit}{bold}{\encodingdefault}{\sfdefault}{bx}{n}

\usepackage{hyperref}
\usepackage{url}
\usepackage{mmstyle}
\usepackage{booktabs}
\usepackage{multirow}
\usepackage{appendix}
\usepackage{graphicx}
\usepackage{marvosym}
\usepackage{ulem}

\makeatletter
\def\thanks#1{\protected@xdef\@thanks{\@thanks
        \protect\footnotetext{#1}}}
\makeatother

\title{Voxurf: Voxel-based Efficient and Accurate Neural Surface Reconstruction}

\author{Tong Wu$^{2}$,
Jiaqi Wang$^1{\textsuperscript{\Letter}}$\thanks{\textsuperscript{\Letter}Corresponding authors. Our code is available at \url{https://github.com/wutong16/Voxurf}.},
Xingang Pan$^{3}$,
Xudong Xu$^2$,
Christian Theobalt$^{3}$,
Ziwei Liu$^4$,
Dahua Lin$^{1,2,5}{\textsuperscript{\Letter}}$ \\
$^1$Shanghai AI Laboratory,
$^2$The Chinese University of Hong Kong,
$^3$Max Planck Institute for Informatics,\\
$^4$S-Lab, Nanyang Technological University,
$^5$Centre of Perceptual and Interactive Intelligence \\
\tt\small
\{wt020, xx018, dhlin\}@ie.cuhk.edu.hk,
wangjiaqi@pjlab.org.cn, \\
\tt\small
\{xpan,theobalt\}@mpi-inf.mpg.de,
ziwei.liu@ntu.edu.sg
}

\newcommand{\revision}[1]{{#1}} 

\iclrfinalcopy 
\begin{document}

\maketitle

\begin{abstract}

Neural surface reconstruction aims to reconstruct accurate 3D surfaces based on multi-view images. Previous methods based on neural volume rendering mostly train a fully implicit model with MLPs, which typically require hours of training for a single scene. 
Recent efforts explore the explicit volumetric representation to accelerate the optimization via memorizing significant information with learnable voxel grids. 
However, existing voxel-based methods often struggle in reconstructing fine-grained geometry, even when combined with an SDF-based volume rendering scheme. 
We reveal that this is because 1) the voxel grids tend to break the color-geometry dependency that facilitates fine-geometry learning, and 2) the under-constrained voxel grids lack spatial coherence and are vulnerable to local minima.
In this work, we present \textbf{Voxurf}, a voxel-based surface reconstruction approach that is both efficient and accurate.
Voxurf addresses the aforementioned issues via several key designs, including 1) a two-stage training procedure that attains a coherent coarse shape and recovers fine details successively, 2) a \textit{dual color network} that maintains color-geometry dependency, and 3) a \textit{hierarchical geometry feature} to encourage information propagation across voxels.
Extensive experiments show that Voxurf achieves high efficiency and high quality at the same time. On the DTU benchmark, Voxurf achieves higher reconstruction quality with a 20x training speedup compared to previous fully implicit methods.
\end{abstract}

\section{Introduction}
\label{sec:introduction}
Neural surface reconstruction based on multi-view images has recently seen dramatic progress. Inspired by the success of Neural Radiance Fields (NeRF)~\citep{mildenhall2020nerf} on Novel View Synthesis (NVS), recent works follow the neural volume rendering scheme to represent the 3D geometry with a signed distance function (SDF) or occupancy field via a fully implicit model~\citep{oechsle2021unisurf,yariv2021volume,wang2021neus}. 
These approaches train a deep multilayer perceptron (MLP), which takes in hundreds of sampled points on each camera ray and outputs the corresponding color and geometry information. 
Pixel-wise supervision is then applied by measuring the difference between the accumulated color on each ray and the ground truth.
Struggling with learning all the geometric and color details with a pure MLP-based framework, these methods require hours of training for a single scene, which substantially limits their real-world applications.

\begin{figure}[t]
	\centering
	\includegraphics[width=1.0\linewidth]{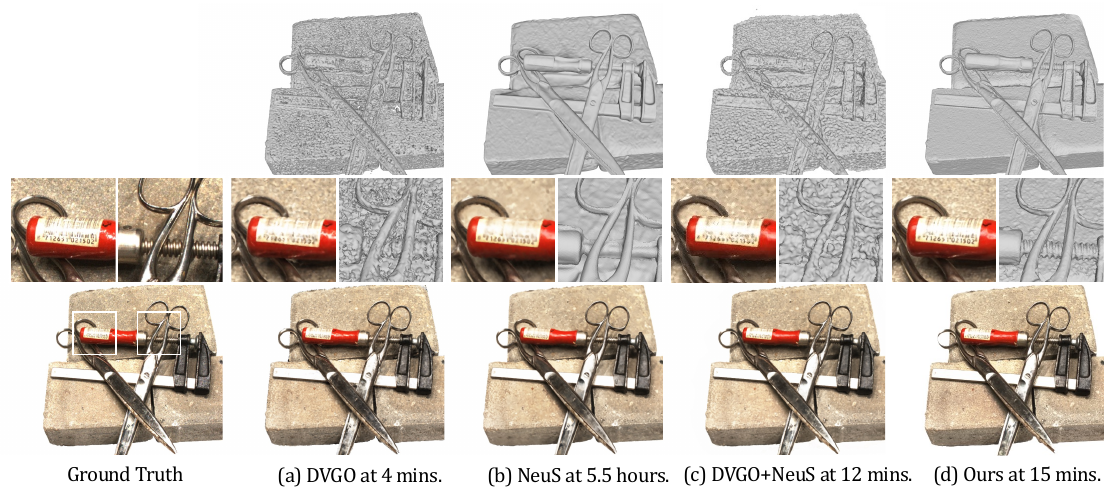}
	\vspace{-25pt}
	\caption{\small
	\textbf{Comparisons among different methods for surface reconstruction and novel view synthesis.} \textbf{(a)} DVGO (v2)~\citep{sun2021direct, sun2022improved} benefits from the fastest convergence but suffers from a poor surface; \textbf{(b)} NeuS~\citep{wang2021neus} produces decent surfaces after a long training time, while high-frequency details are lost in both the geometry and the image; \textbf{(c)} the straightforward combination of DVGO and NeuS produces continuous but noisy surfaces; \textbf{(d)} our method achieves around 20x speedup than NeuS and recovers high-quality surfaces and images with fine details. All the training times are tested on a single Nvidia A100 GPU. 
	}
	\vspace{-17pt}
	\label{fig:teaser}
\end{figure}

Recent advances in NeRF accelerate the training process with the aid of an explicit volumetric representation~\citep{sun2021direct,yu2021plenoxels,chen2022tensorf}. These works directly store and optimize the geometry and color information via explicit voxel grids. For example, the density of a queried point can be readily interpolated from the eight neighboring points, and the view-dependent color is either represented with spherical harmonic coefficients~\citep{yu2021plenoxels} or predicted by shallow MLPs that take learnable grid features as auxiliary inputs~\citep{sun2021direct}. 
These approaches achieve competitive rendering performance at a much lower training cost ($\textless$ 20 minutes).
However, their 3D surface reconstruction results cannot faithfully represent the exact geometry, suffering from conspicuous noise and holes (Fig.~\ref{fig:teaser} (a)). 
It is due to the inherent ambiguity of the density-based volume rendering scheme, and the explicit volumetric representation introduces additional challenges.

In this work, we aim to take advantage of the explicit volumetric representation for efficient training and propose customized designs to harvest high-quality surface reconstruction.
A straightforward idea for this purpose is to embed the SDF-based volume rendering scheme~\citep{wang2021neus,yariv2021volume} into explicit volumetric representation frameworks~\citep{sun2021direct}.
However, we find this na\"{\i}ve baseline model not working well by losing most of the geometry details and producing undesired noise (Fig.~\ref{fig:teaser} (c)).
%
We reveal several critical issues for this framework as follows.
First, in fully implicit models, the color network takes surface normals as inputs, effectively building \textit{color-geometry dependency} that facilitates fine-geometry learning. However, in the baseline model, the color network tends to depend more on the additional \revision{under-constrained} voxel feature grid input, thus breaking color-geometry dependency.
%
Second, due to a high degree of freedom in optimizing a voxel grid, it is hard to maintain a globally coherent shape without additional constraints. Individual optimization for each voxel point hinders the information sharing across the voxel grid, which hurts the surface smoothness and introduces local minima.
We'll unveil these effects and introduce the insight for our architecture design via an empirical study in Sec.~\ref{sec:study}



To tackle the challenges, we introduce \textbf{Voxurf}, an efficient pipeline for accurate \textbf{Vox}el-based s\textbf{urf}ace reconstruction:
1) We leverage a two-stage training process that attains a coherent coarse shape and recovers fine details successively.
2) We design a \textit{dual color network} that is capable of representing a complex color field via a voxel grid and preserving the color-geometry dependency with two sub-networks that work in synergy.
3) We also propose a \textit{hierarchical geometry feature} based on the SDF voxel grid to encourage information sharing in a larger region for stable optimization.
4) We introduce several effective regularization terms to boost smoothness and reduce noise.

We conduct experiments on the DTU~\citep{jensen2014large} and BlendedMVS~\citep{yao2020blendedmvs} datasets for quantitative and qualitative evaluations. 
Experimental results demonstrate that Voxurf achieves lower Chamfer Distance on the DTU~\citep{jensen2014large} benchmark than a competitive fully implicit method NeuS~\citep{wang2021neus} with around 20x speedup. It also achieves remarkable results on the auxiliary task of NVS.  
%
As illustrated in Fig.~\ref{fig:teaser},
our method is shown to be superior in preserving high-frequency details in both geometry reconstruction and image rendering compared to the previous approaches.
In summary, our contributions are highlighted below:
\begin{enumerate}
\item Our approach enables around 20x speedup for training compared to the SOTA methods, reducing the training time from over 5 hours to 15 minutes on a single Nvidia A100 GPU. 
\item Our approach achieves higher surface reconstruction fidelity and novel view synthesis quality, which is superior in representing fine details for both surface recovery and image rendering compared to previous methods. 
\item Our study provides insightful observations and analysis of the architecture design of the explicit volumetric representation framework for surface reconstruction.
\end{enumerate}

\section{Related Works}
\label{sec:related}

\paragraph{\textbf{Multi-view 3D reconstruction}}
Recently, implicit representations that encode the geometry and appearance of a 3D scene by neural networks have gained attention~\citep{park2019deepsdf,chen2019learning,lombardi2019neural,mescheder2019occupancy,sitzmann2019srns,saito2019pifu,atzmon2019controlling,jiang2020sdfdiff,zhang2021learning,toussaint2022hal}. Among them, a plethora of papers have explored neural surface reconstruction from multi-view images.
Methods based on surface rendering \citep{niemeyer2020differentiable,yariv2020multiview,liu2020dist,kellnhofer2021neural} regard the color of an intersection point of the ray and the surface as the final rendered color, while requiring accurate object masks and careful weight initialization.
Recent approaches~\citep{wang2021neus,yariv2021volume,oechsle2021unisurf,francois2021warping,zhang2022nerfusion,liu2020neural,sitzmann2019deepvoxels} based on volume rendering~\citep{max1995optical} formulate the radiance fields and implicit surface representations in a unified model, thereby achieving the merits of both techniques.
However, encoding the whole scene in pure MLP networks requires a long training time. In a departure from these works, we leverage learnable voxel grids and shallow color networks for quick convergence, as well as pursue fine details in surfaces and rendered images. 

\vspace{-7pt}
\paragraph{\textbf{Explicit volumetric representation}}
Despite the success of implicit neural representations in 3D modeling, recent advances have integrated explicit 3D representations, \eg, point clouds, voxels, and MPIs~\citep{mildenhall2019llff}, and received growing attention~\citep{wizadwongsa2021nex, xu2022point,lombardi2019neural,wang2022gosurf,fang2022tineuvox}.
%
%
%
%
Instant-ngp~\citep{mueller2022instant} uses multi-resolution hashing for efficient encoding and implements fully-fused CUDA kernels.
Plenoxels~\citep{yu2021plenoxels} represent a scene as a sparse 3D grid with spherical harmonics and are optimized two orders of magnitude faster than NeRF~\citep{mildenhall2020nerf}.
TensoRF~\citep{chen2022tensorf} considers the full volume field as a 4D tensor and factorizes it into multiple compact low-rank tensor components for efficiency.
The method most related to ours is DVGO~\citep{sun2021direct}, which adopts a hybrid architecture design including voxel grids and a shallow MLP.
Despite their remarkable results on novel view synthesis, none of them is designed to faithfully reconstruct the geometry of the scene.
In contrast, we target at not only rendering photo-realistic images from novel viewpoints but also reconstructing high-quality surfaces with fine details.
\section{Preliminaries}
\label{sec:preliminaries}
%

\noindent\textbf{{Volume rendering with SDF representation.}}
NeuS~\citep{wang2021neus} represents a scene as an implicit SDF field parameterized by an MLP. The ray emitting from the camera center $o$ through an image pixel in the viewing direction $v$ can be expressed as $\{p(t) = o + tv | t \geq 0\}$. The rendered color for the image pixel is integrated along the ray with volume rendering~\citep{max1995optical}, which is approximated by $N$ discrete sampled points $\{p_i = o + t_iv | i=1, ..., N, t_i < t_{i+1} \}$ on the ray:
\begin{equation}
    \hat{C}(r) = \sum_{i=1}^{N}T_i \alpha_i c_i, \ \ T_i = \prod \limits_{j=1}^{i-1} (1-\alpha_j),
    \label{eq:volum_render}
\end{equation}
where $\alpha_i$ is the opacity value, and $T_i$ is the accumulated transmittance.
The key difference between NeuS and NeRF is the formula of $\alpha_i$. 
In NeuS, $\alpha_i$ is formulated as:
\begin{equation}
    \alpha_i = \max\left(\frac{\Phi_s(f(p(t_i))) - \Phi_s(f(p(t_{i+1})))}{\Phi_s(f(p(t_i)))}, 0\right).
    \label{eq:neus_alpha}
\end{equation}
Here, $f(x)$ is the SDF function, and $\Phi_s(x) = (1 + e^{-sx})^{-1}$ is the Sigmoid function, where the $s$ value is learned or manually updated during training.

\noindent\textbf{{Explicit volumetric representation.}}
DVGO~\citep{sun2021direct} represents the geometry with explicit density voxel grids $V^{(density)} \in \mathbb{R}^{1 \times N_x \times N_y \times N_z}$. It applies a hybrid architecture for color prediction that comprises a shallow MLP parameterized by $\Theta$ and a feature voxel grid $V^{(feat)} \in \mathbb{R}^{C \times N_x \times N_y \times N_z}$.
Given a 3D position $p$ and the viewing direction $v$, the volume density $\sigma$ and color $c$ are estimated with:
\begin{equation}
    \sigma = {\rm interp}(p, V^{(density)}),
\end{equation}
\begin{equation}
    c = {\rm MLP}_{\Theta}({\rm interp}(p, V^{(feat)}), p, v),
    \label{eq:color_dvgo}
\end{equation}
where `interp' denotes the trilinear interpolation.
Following NeRF~\citep{mildenhall2020nerf,tancik2020fourier}, the positional encoding for both $p$ and $v$ is applied in Eqn.~\ref{eq:color_dvgo}.

\noindent\textbf{{Na\"{\i}ve Combination.}}
A straightforward combination of the two techniques is to replace the volume rendering in DVGO with the SDF-based volume rendering scheme as in Eqn.~\ref{eq:volum_render} and Eqn.~\ref{eq:neus_alpha}.
It serves as the na\"{\i}ve baseline in this work, which can hardly produce satisfactory results, as shown in Fig.~\ref{fig:teaser} (c). 
We will cast light on this phenomenon via an empirical study in the next section.

\section{Study on architecture design for geometry learning}
\label{sec:study}
\begin{figure}[t]
	\centering
	\includegraphics[width=1.0\linewidth]{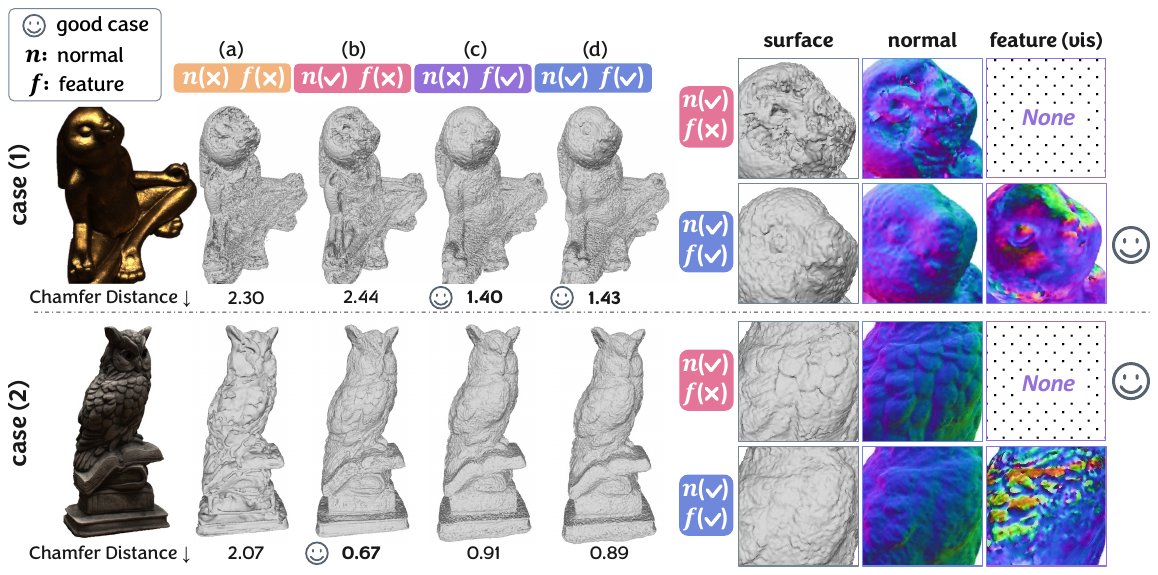}
	\vspace{-25pt}
	\caption{\small
	\textbf{Reconstruction results from different architecture designs.} The \textit{surface normal} $n$ and \textit{learnable feature} $f$ are both optional inputs to the color network.
	We show results of two cases under four settings on the \textbf{left}, and we zoom in to analyze the surfaces, normal fields, and feature fields on the \textbf{right}. Case (1) (a, c) and (b, d) show that the feature $f$ helps maintain a coherent shape, while case (2) (b, d) reveal that it discourages the reconstruction of geometry details since it disturbs the color-geometry dependency built by the normal $n$.}
	\vspace{-17pt}
	\label{fig:study}
\end{figure}

In this section, we carry out some prior experiments with variants of the baseline model, aiming to figure out the key factors for architecture design in this task. 
Specifically, we employ an SDF voxel grid $V^{(sdf)}$ and apply Eqn.~\ref{eq:neus_alpha} for $\alpha$ calculation with a manually defined schedule for $s$.
We start with a shallow MLP as the color network, where 1) the local feature $f$ interpolated from $V^{(feat)}$ and 2) the normal vector $n$ calculated by $V^{(sdf)}$ are both optional inputs.
A decent surface reconstruction is expected to possess \textit{a coherent coarse structure}, \textit{accurate fine details}, and \textit{a smooth surface}. We will next focus on these factors and analyze the effects of different architecture designs. 

\noindent{\textbf{The key to maintaining a coherent coarse shape.}}
Intuitively, the capacity of a shallow MLP is limited, and it can hardly represent a complex scene with different materials, high-frequency textures, and view-dependent lighting information. 
When the ground truth image encounters a rapid color-shifting, the volume rendering integration over an under-fitted color field results in a corrupted geometry, as shown in Fig.~\ref{fig:study} case (1) (a) and (b).
Incorporating the local feature $f$ enables fast color learning and increases the representation ability of the network, and the problem is noticeably alleviated, as shown in Fig.~\ref{fig:study} case (1), the differences between (a) and (c), (b) and (d).

\noindent{\textbf{The key to reconstructing accurate geometry details.}}
We then introduce another case in Fig.~\ref{fig:study} case (2). 
Its texture changes moderately, and the color is largely correlated with the surface normal due to diffuse reflection.
Although the geometry still collapses given neither normal $n$ or feature $f$ as input in Fig.~\ref{fig:study} case (2) (a), we can observe a reasonable reconstruction even with some geometry details in Fig.~\ref{fig:study} case (2) (b) with only $n$ as the input. Incorporating the feature $f$ does not further reduce the Chamfer Distance (CD); instead, geometry details are missing since the learnable feature $f$ disturbs the geometry-color dependency, \ie, the relationship built between the color and the surface normal, as shown in Fig.~\ref{fig:study} case (2), the differences between (b) and (d). \revision{Considering the cons and pros of the learnable local feature, it is valuable to design a framework that leverages it for coherent shape and keeps color-geometry dependency for fine details.}

\noindent{\textbf{The reason for noisy surfaces.}} For all the cases above, the results suffer from obvious noise on the surface. Compared with learning an implicit representation globally, the under-constrained voxel grids lack spatial coherence and are vulnerable to local minima, which hurts the continuity and smoothness of the surface. \revision{An intuitive idea is leveraging geometry cues from a region rather than a local point, which can be introduced in model inputs, network components, and loss functions. }



\section{Methodology}
\label{sec:method}
Inspired by the insight revealed in Sec.~\ref{sec:study}, we propose several key designs:
1) we adopt a two-stage training procedure that attains a coherent coarse shape (Sec.~\ref{subsec:coarse}) and recovers fine details (Sec.~\ref{subsec:fine}) successively;
2) we propose a dual color network to maintain color-geometry dependency and recover precise surfaces and novel-view images;
3) we design a hierarchical geometry feature to encourage information propagation across voxels for stable optimization;
4) we also introduce smoothness priors, including a gradient smoothness loss for better visual quality (Sec.~\ref{subsec:tv}).

\vspace{-5pt}
\subsection{Coarse shape initialization}
\label{subsec:coarse}
%

We initialize our SDF voxel grid $V^{(sdf)}$ with an ellipsoid-like zero level set inside a prepared region for reconstruction as in \citep{sun2021direct}. We then perform coarse shape optimization with the aid of $V^{(feat)}$ as introduced in Sec.~\ref{sec:study}.
Specifically, we train a shallow MLP with both normal vector $n$ and local feature $f$ as inputs, along with the embedded position $p$ and viewing direction $v$. 
To encourage a stable training process and smooth surface, we propose to conduct the interpolation on a smoothed voxel grid rather than the raw data of $V^{(sdf)}$. 
In particular, we denote $\mathcal{G}(V,k_g,\sigma_g)$ as applying 3D convolution on the voxel grid $V$ with a Gaussian kernel, whose weight matrix follows a Gaussian distribution:
$K_{i,j,k} = 1/Z \times \rm exp \left(-((i-\lfloor k_g / 2 \rfloor)^2 + (j-\lfloor k_g / 2 \rfloor)^2 + (k- \lfloor k_g / 2 \rfloor)^2) / 2\sigma_g^2\right), i,j,k \in\{0,1,..., k_g-1\}$,
where $Z$ denotes a normalization term, $k_g$ denotes the kernel size, and $\sigma_g$ denotes the standard deviation.
Querying a smoothed SDF value $d'$ of an arbitrary point $p$ thus becomes:
\begin{equation}
    d' = {\rm interp}(p, \mathcal{G}(V^{(sdf)},k_g,\sigma_g)).
\end{equation}
We use $d'$ for the ray marching integration following Eqn.~\ref{eq:volum_render} and Eqn.~\ref{eq:neus_alpha} and calculate the reconstruction loss.
We also apply several smoothness priors as to be introduced in Sec.~\ref{subsec:tv}

\subsection{Fine geometry optimization}
\label{subsec:fine}
\begin{figure}[t]
	\centering
	\includegraphics[width=1.0\linewidth]{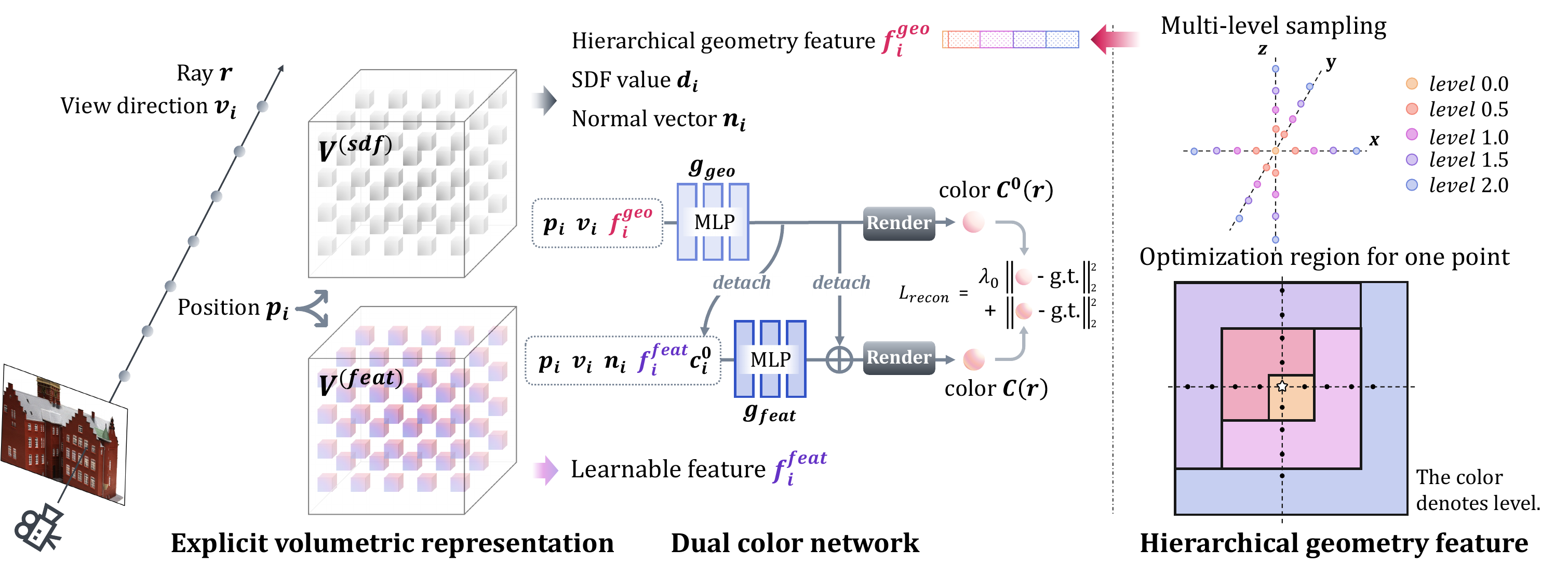}
	\vspace{-20pt}
	\caption{\small
	\textbf{Overview of key components in our model.} We adopt an explicit volumetric representation with an SDF voxel grid $V^{(sdf)}$ and a feature voxel grid $V^{(feat)}$. In the middle, we show the design for our dual color network, where $f^{feat}_i$ is the interpolated feature from $V^{(feat)}$ at point $p_i$, and $f^{geo}_i$ denotes the hierarchical feature constructed on the right. Here we show the multi-level sampling scheme and the region of grids that is affected by one point during optimization with different settings of levels.
	}
	\vspace{-15pt}
	\label{fig:pipeline}
\end{figure}

At this stage, we aim to recover accurate geometry details based on the coarse initialization.
We note that the challenges are two-fold:
\textbf{1)} The study in Sec.~\ref{sec:study} reveals a trade-off introduced by the feature voxel grid, \ie, the representation capacity of the color field is improved at the sacrifice of color-geometry dependency.
\textbf{2)} The optimization of the SDF voxel grid is based on trilinear interpolation to query a 3D point. The operation brings in fast convergence, while it also limits information sharing across different locations, which may lead to local minima with degenerate solutions and a sub-optimal smoothness.
We propose a \textit{dual color network} and a \textit{hierarchical geometry feature} to address these two issues, respectively. 

\noindent{\textbf{Dual color network.}}
The observation in Sec.~\ref{sec:study} encourages us to design a dual color network that takes advantage of the local feature $f_i^{feat}$ interpolated from the learnable feature voxel grid $V^{(feat)}$ without losing the color-geometry dependency.
As shown in Fig.~\ref{fig:pipeline}, we train two shallow MLPs with different additional inputs besides the embedded position and view direction. 
The first MLP $g_{geo}$ takes the hierarchical geometry feature $f_i^{geo}$, which will be introduced later, to build the color-geometry dependency; 
the second one $g_{feat}$ takes both a simple geometry feature (\ie, the surface normal $n_i$) and the local feature $f_i^{feat}$ as inputs to enable a faster and more precise color learning, which will in turn benefit the geometry optimization.
The two networks are combined in a residual manner with detaching operations: the output of $g_{geo}$, denoted by $c_0$, is detached before input to $g_{feat}$, and the output is added back to a detached copy of $c_0$ to get the final color prediction $c$. 

Outputs of both $g_{geo}$ and $g_{feat}$ are supervised by a reconstruction loss between the ground truth image and the integrated color along the ray. 
Specifically, the rendered colors from them are denoted as $C^0(r)$ and $C(r)$, and the overall reconstruction loss is formulated as:
\begin{equation}
    \mathcal{L}_{recon} = \frac{1}{\mathcal{R}}\sum\limits_{r\in\mathcal{R}}\left(||C(r)-\hat{C}(r)||^2_2 + \lambda_0||C_0(r)-\hat{C}(r)||^2_2\right),
\end{equation}
where $\hat{C}(r)$ denotes the ground truth color, and $\lambda_0$ denotes a loss weight. 
$V^{(feat)}$ and the MLP $g_{feat}$ fit the scene field rapidly, while the MLP $g_{geo}$ fits the scene at a relatively slower pace. The detaching operations promote a stable optimization of $g_{geo}$ guided by the reconstruction loss of itself, which helps preserve color-geometry dependency.



\noindent{\textbf{Hierarchical geometry feature.}}
Using the surface normal $n$ as the geometry feature for the color networks is a straightforward choice, while it takes in information only from adjacent grids of $V^{(SDF)}$. 
In order to enlarge the perception area and encourage information propagation across voxels, we propose to look at a larger region of the SDF field and take the corresponding SDF values and gradients as an auxiliary condition to the color networks. 
Specifically, for a given 3D position $p = (x, y, z)$, we take half of the voxel size $v_s$ as the step size and define its neighbours along the $X, Y, Z$ axis on both sides. Taking the $X$ axis as an example, the neighbouring coordinates are defined as $p_x^{l-} = (x^{l-}, y, z)$ and $p_x^{l+}=(x^{l+}, y, z)$, where $x^{l-} = \max(x - l*v_s, 0)$, $x^{l+} = \min(x + l*v_s, v_x^m)$, $l \in {[0.5, 1.0, 1.5, ...]}$ denotes the `level' of neighbour area, and $v_x^m$ denotes the maximum of the voxel grid on $x$ axis. 
We then extend the definition to a hierarchical manner by concatenating the neighbours from different levels together as formulated below:
\begin{equation}
\begin{split}
    d_k^l = [d_k^{l-}, d_k^{l+}] = [{\rm interp}(p_k^{l-}, V^{(sdf)}), {\rm interp}(p_k^{l+}, V^{(sdf)})], k \in \{x, y, z\}, \\
    f^{sdf}_p(l) = [d^{0}, d_x^{0.5}, d_y^{0.5}, d_z^{0.5}, \cdot\cdot\cdot, d_x^{l}, d_y^{l}, d_z^{l}]^T, \quad \quad\quad
    \label{eq:feat_sdf}
\end{split}
\end{equation}
where $d_x^{l}$ denotes the SDF values queried from $V^{(sdf)}$ at locations $p_x^{l-}$ and $p_x^{l+}$. When $l=0$, $f^{sdf}_p(0) = d^{0}$, which is exactly the SDF value at the location $p$ itself.
Then, we also incorporate the gradient information into the geometry feature. Specifically, we can \revision{obtain} the gradient vector $\delta^l_x = (d_x^{l+} - d_x^{l-}) / (2 * l * v_s)$. \revision{We normalize the $[\delta^l_x, \delta^l_y, \delta^l_z]$ to a L2-norm of 1, denoted as $n^l \in \mathbb{R}^3$.}
The hierarchical version of the normal is formulated as:
\begin{equation}
    f^{normal}_p(l) = [n^{0.5}, \cdot\cdot\cdot, n^{l}].
    \label{eq:feat_normal}
\end{equation}
Finally, the hierarchical geometry feature at point $p$ for a predefined level $l \in [0.5, 1.0, 1.5, ...]$ is to combine the information above by:
\begin{equation}
    f^{geo}_p(l) = [f^{sdf}_p(l), f^{normal}_p(l)].
    \label{eq:feat_geo}
\end{equation}
As shown in Fig.~\ref{fig:pipeline}, $f^{geo}_p(l)$ is input to the MLP $g_{geo}$ to assist geometry learning. 

\subsection{Smoothness priors}
\label{subsec:tv}
We incorporate \revision{two} effective regularization terms to facilitate surface smoothness during training. \textbf{(1)} First, we adopt a total variation (TV) regularization~\citep{rudin1994total}:
\begin{equation}
    \mathcal{L}_{TV}(V) = \sum \limits_{d \in [D]}\sqrt{\Delta_x^2(V,d) + \Delta_y^2(V,d) + \Delta_z^2(V,d)},
    \vspace{-5pt}
\end{equation}
where $\Delta_x^2(V,d)$ denotes the squared difference between the value of $d$th channel in voxel $v := (i; j; k)$ and the $d$th value in voxel $(i + 1; j; k)$, which can be analogously extended to $\Delta_y^2(V,d)$ and $\Delta_z^2(V,d)$.
We apply the TV term above to the SDF voxel grid, denoted by $\mathcal{L}_{TV}(V^{(sdf)})$, which encourages a continuous and compact geometry.

\textbf{(2)} We also assume the surface to be smooth in a local area, and we follow the definition of the Gaussian convolution in Sec.~\ref{subsec:coarse} and introduce a smoothness regularization formulated as:
\begin{equation}
    \mathcal{L}_{smooth}(V) = || \mathcal{G}(V, k_g, \sigma_g) - V||^2_2,
    \label{eq:smooth}
\end{equation}
We apply the smoothness term above to the gradient of SDF voxel grid for a \textit{gradient smoothness loss}, denoted by $\mathcal{L}_{smooth}(\nabla V^{(sdf)})$. It encourages a smooth surface and alleviates the issue of noisy points in the free space.
Notice that we can also naturally conduct post-processing on the SDF field after training, thanks to its explicit representation. For example, applying the Gaussian kernel above before extracting the geometry can further boost surface smoothness for better visualization.

Finally, the overall training loss is formulated as:
\begin{equation}
    \mathcal{L} = \mathcal{L}_{recon} + \lambda_{tv}\mathcal{L}_{TV}(V^{(sdf)}) + \lambda_s\mathcal{L}_{smooth}(\nabla V^{(sdf)}),
    \label{eq:reg}
\end{equation}
where $\lambda_{tv}$ and $\lambda_s$ denote the weights for the corresponding loss terms.
 
\section{Experiments}
\label{sec:expriments}

\noindent{\textbf{Experimental setup.}}
We use the DTU~\citep{jensen2014large} dataset for quantitative and qualitative comparisons and show qualitative results on several challenging scenes from the BlendedMVS~\citep{yao2020blendedmvs} dataset.
We include several baselines for comparisons:
1) IDR~\citep{yariv2020multiview},
2) NeuS~\citep{wang2021neus},
3) NeRF~\citep{mildenhall2020nerf},
4) DVGO~\citep{sun2021direct},
5) Point-NeRF~\citep{xu2022point}.
The results of 1), 2), and 3) are taken from the original papers~\citep{yariv2020multiview,wang2021neus}; for 5), we use the neural point for evaluation. We provide a clean background for all the methods for a fair comparison. 
Experimental results with non-empty backgrounds and comparisons with more methods~\citep{schonberger2016pixelwise, oechsle2021unisurf, yariv2021volume} are included in the supplementary materials. 
Please also refer to the supplementary materials for further descriptions of the datasets, baseline methods, and implementation details.

%
\begin{table}[t]
  \centering
  \scriptsize
  \caption{\small Quantitative evaluation on DTU dataset.}
   \setlength\tabcolsep{3.2pt}
    \begin{tabular}{l|cccccccccccccccc}
    \toprule
    \toprule
    Scan & 24    & 37    & 40    & 55    & 63    & 65    & 69    & 83    & 97    & 105   & 106   & 110   & 114   & 118   & 122   & mean \\
    \midrule
    NeRF\citep{mildenhall2020nerf} & 1.83  & 2.39  & 1.79  & 0.66  & 1.79  & 1.44  & 1.50  & 1.20  & 1.96  & 1.27  & 1.44  & 2.61  & 1.04  & 1.13  & 0.99  & 1.54 \\
    IDR\citep{yariv2020multiview} & 1.63  & 1.87  & 0.63  & 0.48  & 1.04  & 0.79  & 0.77  & 1.33  & 1.16  & 0.76  & 0.67  & 0.90  & 0.42  & 0.51  & 0.53  & 0.90 \\
    DVGO\citep{sun2021direct} & 1.83  & 1.74  & 1.70  & 1.53  & 1.91  & 1.91  & 1.77  & 2.60  & 2.08  & 1.79  & 1.76  & 2.12  & 1.60  & 1.80  & 1.58  & 1.85 \\
    PointNeRF\citep{xu2022point} & 0.87 & 2.06 & 1.20 & 1.01 & 1.01 & 1.39 & 0.80 & \textbf{1.04} & \textbf{0.92} & 0.74 & 0.97 & \textbf{0.76} & 0.56 & 0.90 & 1.05 & 1.02 \\
    NeuS\citep{wang2021neus} & 0.83  & 0.98  & 0.56  & 0.37  & 1.13  & \textbf{0.59}  & \textbf{0.60}  & 1.45  & 0.95  & 0.78  & \textbf{0.52}  & 1.43  & \textbf{0.36}  & \textbf{0.45}  & \textbf{0.45}  & 0.77 \\
    DVGO + NeuS & 1.24  & 0.87  & 0.74  & 0.48  & 1.20  & 1.41  & 1.113  & 1.96  & 1.44  & 0.98  & 1.13  & 1.99  & 1.62  & 0.77  & 0.62  & 1.13 \\
    Ours  & \textbf{0.65}  & \textbf{0.74}  & \textbf{0.39}  & \textbf{0.35}  & \textbf{0.96}  & 0.64  & 0.85  & 1.58  & 1.01  & \textbf{0.68}  & 0.60  & 1.11  & 0.37  & \textbf{0.45}  & 0.47  & \textbf{0.72} \\
    \bottomrule
    \bottomrule
    \end{tabular}%
    \vspace{-15pt}
  \label{tab:dtu_main}%
\end{table}%

\begin{figure}[t]
	\centering
	\includegraphics[width=0.95\linewidth]{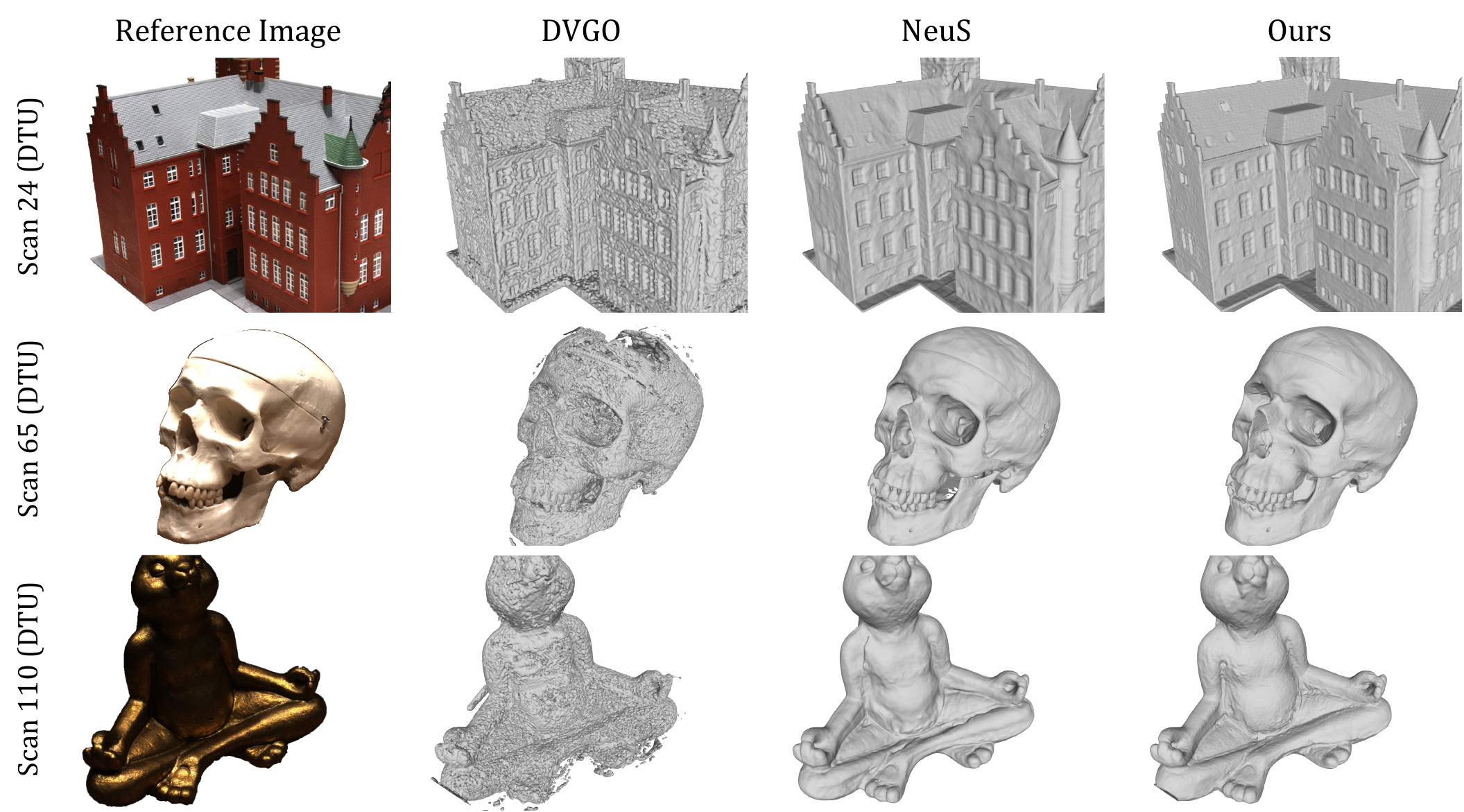}
	\vspace{-10pt}
	\caption{\small
	Qualitative comparisons on the DTU dataset. See more scenes in supplementary materials.
	}
	\vspace{-10pt}
	\label{fig:dtu}
\end{figure}

\begin{figure}[t]
	\centering
	\includegraphics[width=0.95\linewidth]{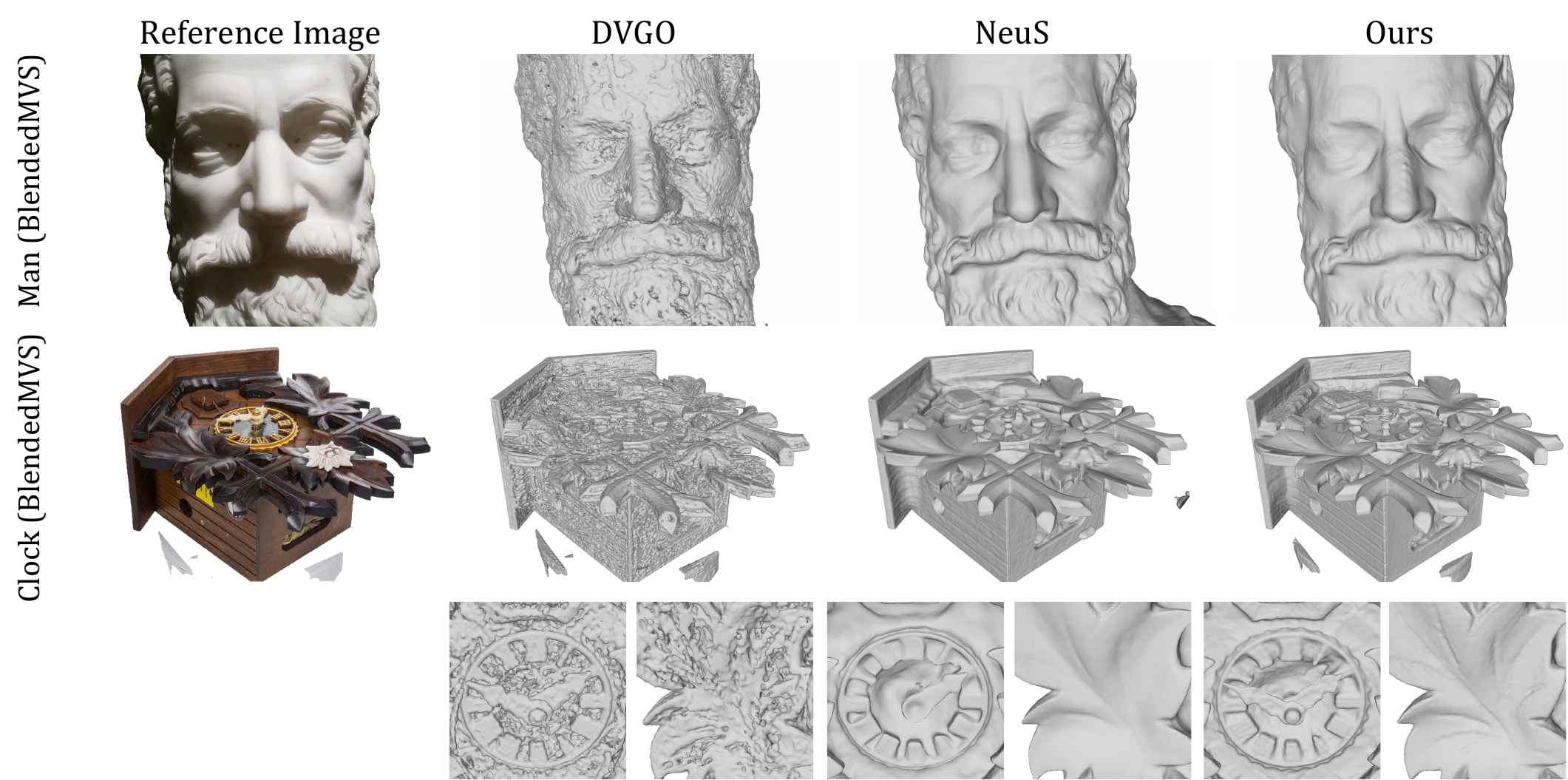}
	\vspace{-5pt}
	\caption{\small
	Qualitative comparisons on the BlendedMVS dataset. See more scenes in supplementary materials.
	}
	\vspace{-15pt}
	\label{fig:bmvs}
\end{figure}

\begin{table}[t]
  \centering
  \scriptsize
  \caption{\small An overall comparison on surface reconstruction, novel view synthesis, and training time on DTU. 
  }
    \begin{tabular}{l|cccc|c}
    \toprule
    \toprule
          & PSNR $\uparrow$ & SSIM $\uparrow$  & LPIPS $\downarrow$ & CD $\downarrow$   & Time (Nvidia A100) \\
    \midrule
    DVGO~\citep{sun2021direct}  & 31.64 & 0.916 & 0.159 & 1.85  & 4 mins \\
    NeuS~\citep{wang2021neus}  & 29.63 & 0.892 & 0.199 & 0.77  & 5.5 hours \\
    Ours  & \textbf{32.16} & \textbf{0.929} & \textbf{0.144} & \textbf{0.72} & 15 mins \\
    \bottomrule
    \bottomrule
    \end{tabular}%
    \vspace{-10pt}
  \label{tab:overall}%
\end{table}%

\subsection{Comparisons}
%
The quantitative results for surface reconstruction on DTU are reported in Table~\ref{tab:dtu_main}. 
Quantitative experimental results show that we achieve lower Chamfer Distances than previous methods under the same setting.
We conduct qualitative comparisons on both DTU and BlendedMVS in Fig.~\ref{fig:dtu} and Fig.~\ref{fig:bmvs}, respectively. DVGO shows poor reconstruction quality with noise and holes since it is designed for novel view synthesis rather than surface reconstruction.
NeuS and ours show accurate and continuous surface recovery in a variety of cases. In comparison, NeuS, as a fully implicit model, naturally benefits from the intrinsic continuity and encourages smoothness in local areas, while it sometimes fails to recover very thin geometry details due to over-smoothing. 
In contrast, our method is superior in recovering fine geometry details thanks to our designs in Sec.~\ref{sec:method}.

We further perform a more extensive evaluation of our method on surface reconstruction, novel view synthesis, and training time in Table~\ref{tab:overall}. Our method outperforms DVGO and NeuS on both surface reconstruction and novel view synthesis by a clear margin on all the metrics. Notably, our method achieves around \textbf{20x} speedup compared to NeuS for producing high-quality surface reconstruction.

\subsection{Analysis}
\label{subsec:analysis}

\begin{figure}[t]
	\centering
	\includegraphics[width=0.98\linewidth]{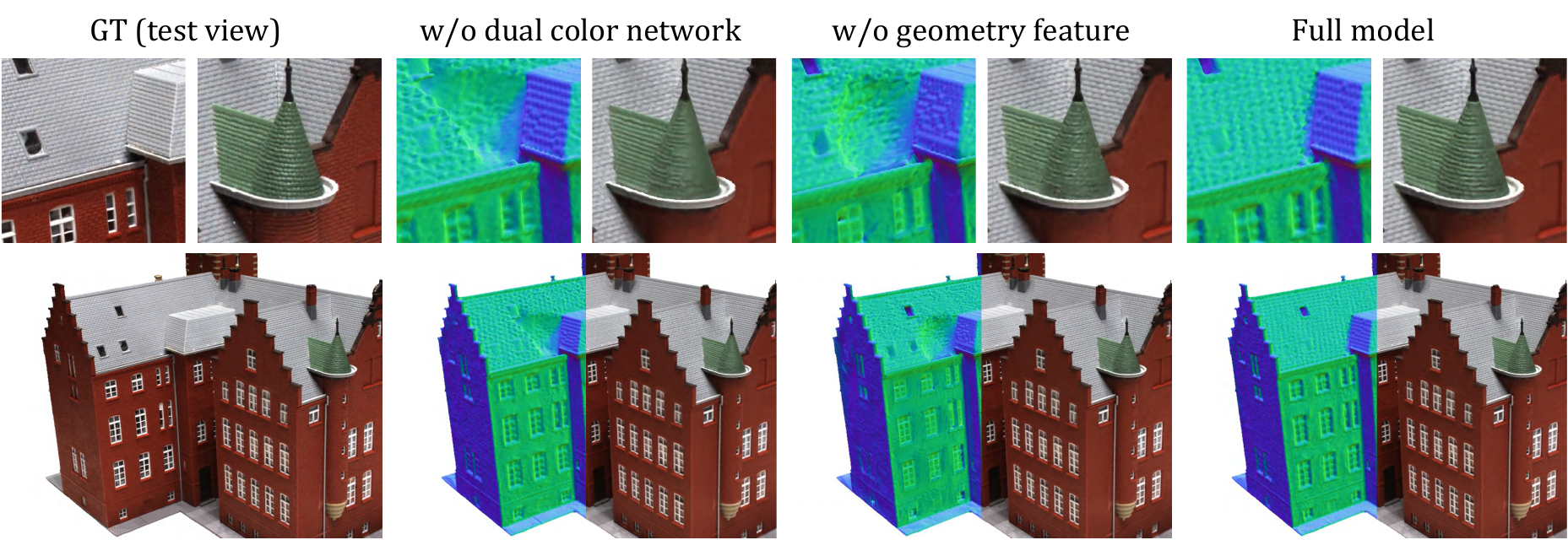}
	\vspace{-5pt}
	\caption{\small
    The dual color network learns the color field for complex scenes well and preserves color-geometry dependency, which facilitates geometry learning (see the roofs); the hierarchical geometry feature promotes accurate surface reconstruction (see the windows).
	}
	\label{fig:ablation}
	\vspace{-15pt}
\end{figure}

\begin{table}[t]
\begin{minipage}[t]{\textwidth}
 \begin{minipage}[t]{0.45\textwidth}
  \centering
     
     \caption{\small Ablation over the effect of dual color network and hierarchical geometry feature.
     }
    \small
    \begin{tabular}{c|cccc}
    \toprule
    \toprule
    \small
    CD  & 0.91  & 0.79  & 0.77  & \textbf{0.72} \\
    \midrule
    Dual  &       & \checkmark  &       & \checkmark \\
    Hierarchical &       &       & \checkmark  & \checkmark \\
    
    \bottomrule
    \bottomrule
    \end{tabular}%
  \label{tab:two_module}%
  \end{minipage}
  \hspace{20pt}
  \begin{minipage}[t]{0.45\textwidth}
  \centering
   
   \caption{\small Ablation over the \textit{Residual} and \textit{Detach} designs of the dual color network (Sec.~\ref{subsec:fine}). 
       }
  \small
    \begin{tabular}{l|cccc}
    \toprule
    \toprule
    \small
    
    CD & 0.77  & 0.75  & 0.75  & \textbf{0.72} \\
    \midrule
    Residual  &       & \checkmark  &       & \checkmark \\
    Detach &       &       & \checkmark  & \checkmark \\
    
    \bottomrule
    \bottomrule
    \end{tabular}%
  \label{tab:dualnet}%
  \end{minipage}
\end{minipage}
\vspace{-10pt}
\end{table}

In this section, we carry out a series of ablation studies to evaluate each technical component.

\noindent{\textbf{The effect of the dual color network and a hierarchical geometry feature.}}
As shown in Table~\ref{tab:two_module}, both techniques individually work well on the baseline model, and a combination of them produces the best result.
\textbf{1)} The effect of dual color network can be directly sensed in the improvement of image rendering quality, as can be seen from the comparison of roof textures in Fig.~\ref{fig:ablation}. An accurate color field and the color-geometry dependency will promote geometry learning, as can be observed from the comparison of roof geometries (viewed in normal images) in Fig.~\ref{fig:ablation}. Experimental results in Table~\ref{tab:dualnet} also validate the effectiveness of the design introduced in Sec.~\ref{subsec:fine}, including the residual color and detachment.
\textbf{2)} Hierarchical geometry feature directly promotes an accurate surface reconstruction, as demonstrated by results in Table~\ref{tab:two_module} and the difference between normal images of Fig.~\ref{fig:ablation}. We also explore different design details, including the level selection and the effects of gradient and SDF value in supplementary materials.

\begin{figure}[h]
	\centering
	\includegraphics[width=0.98\linewidth]{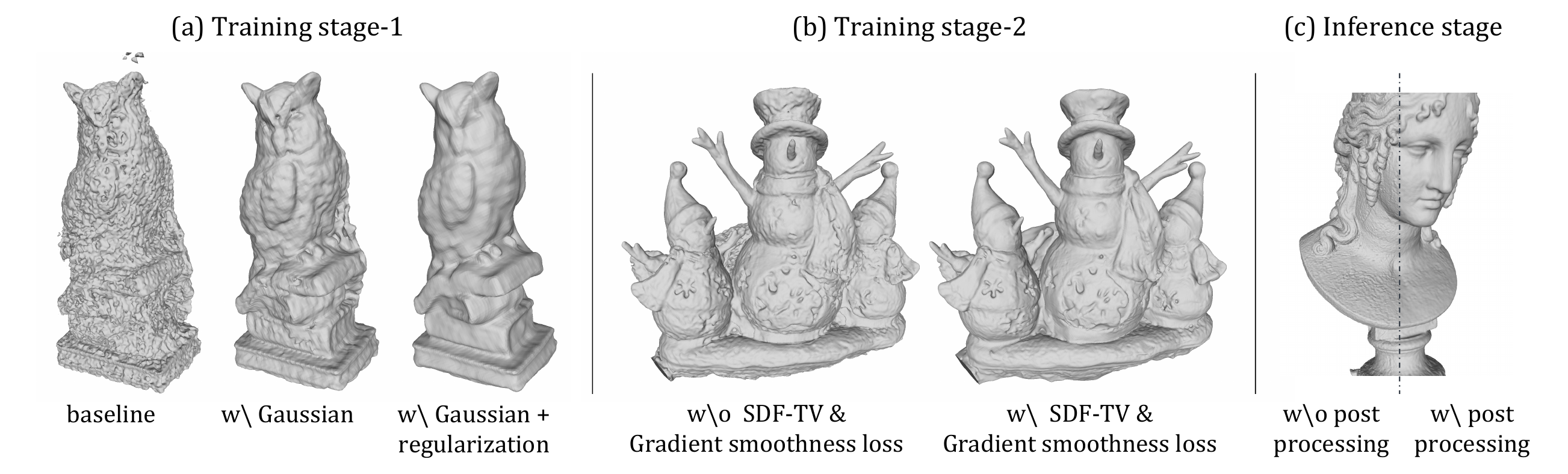}
	\vspace{-15pt}
	\caption{\small
	Studies on technical components that encourage surface smoothness during the (a) coarse shape initialization, (b) fine geometry optimization, and (c) post-processing stage.
	}
	\vspace{-15pt}
	\label{fig:smooth}
\end{figure}

\noindent{\textbf{Ablation over smoothness priors.}}
We make efforts to encourage the continuity and smoothness of the reconstructed surface at different stages. 
As shown in Fig.~\ref{fig:smooth} (a), during the coarse shape initialization stage, the naive solution produces holes and noises. 
Applying the Gaussian convolution substantially alleviates the problem and leads to a more compact geometry. Regularization terms including the TV and our gradient smoothness loss would further encourage a clean and smooth surface to provide a good initialization for the next stage.
Fig.~\ref{fig:smooth} (b) shows that during the fine geometry optimization stage, the regularization terms also help maintain surface smoothness. 
Finally, as shown in Fig.~\ref{fig:smooth} (c), post processing on a trained model can promote surface smoothness for a better visualization quality and maintain an accurate structure at the same time. 
An ablation study on the effects of the SDF TV term and our gradient smoothness loss is in the supplementary materials.

\section{Conclusion}
This paper proposes \textbf{Voxurf}, a voxel-based approach for efficient and accurate neural surface reconstruction. It includes several key designs: the two-stage framework attains a coherent coarse shape and recovers fine details successively; the dual color network helps maintain color-geometry dependency, and the hierarchical geometry feature encourages information propagation across voxels; effective smoothness priors including a gradient smoothness loss further improve the visual quality.
Extensive experiments show that Voxurf achieves high efficiency and high quality at the same time. 

\section{Acknowledgement}
This work is supported by Shanghai AI Laboratory, NTU NAP, MOE AcRF Tier 2 (MOE-T2EP20221-0012), ERC Consolidator Grant 4DRepLy (770784), and under the RIE2020 Industry Alignment Fund – Industry Collaboration Projects (IAF-ICP) Funding Initiative
\appendix
\setcounter{table}{0}
\setcounter{figure}{0}
\renewcommand{\thetable}{R\arabic{table}}
\renewcommand\thefigure{S\arabic{figure}}

\section{Experimental details}
\label{supp:implement}
\subsection{Datasets.}
The \textbf{DTU}~\citep{jensen2014large} dataset contains different static scenes with 49 or 64 posed multi-view images for each scene. It covers a variety of objects with different materials, geometry, and texture. 
We evaluate our approach on DTU with the same 15 scenes following IDR~\citep{yariv2020multiview} and quantitatively compare it with previous work on Chamfer Distance, given the ground truth point clouds.
The \textbf{BlendedMVS}~\citep{yao2020blendedmvs} dataset contains 113 scenes that cover a variety of real-world environments, providing 31 to 143 posed multi-view images for each. We select 7 challenging scenes following NeuS~\citep{wang2021neus} and present qualitative comparisons with previous works.

\subsection{Implementation details}
We set the expected number of voxels to be $96^3$ at the coarse stage and $256^3$ at the fine stage, including an up-scaling step. 
We use a batch size of 8,192 rays with the point sampling step size on a ray to be half of the voxel size. 
We train our coarse initialization stage for 10k iterations and the fine geometry optimization stage for 20k iterations with an Adam optimizer. The initial learning rate is set as $1^{-3}$ for all the MLPs and $0.1$ for voxels in the coarse stage, while the SDF voxel starts by $5^{-3}$ in the fine stage.

We use the same hyper-parameters for all scenes. 
We use a 3-layer MLP for the coarse training stage and two 4-layer MLPs for the dual color network in the fine training stage. We choose level 2.0 for the hierarchical geometry feature and set the dimension of the learnable feature voxel grid as 6.

For the coarse stage, we set $\lambda_{tv}=10^{-4}$ and $\lambda_s=2\times10^{-4}$ for the regularization terms and introduce an additional TV term for $V^{(feat)}$ with a weight of $10^{-2}$. The Gaussian kernel is $5^3$ in size with $\sigma_g=0.8$. 
For the fine stage, we set $\lambda_0=0.5$ for the reconstruction loss; we set $\lambda_{tv}=10^{-3}$ and $\lambda_s=5\times10^{-4}$ for the regularization terms. The fine SDF grid starts with a resolution of $160^3$, which is then up-scaled by trilinear interpolation to $256^3$ after 15000 iterations. 

For the $s$ value in $\Phi_s$ in Eqn. 2, we design a function based on the iteration, $s = 1/(i/r+1/s_{start})$, where $s_{start}$ controls the beginning value of $s$, $i$ denotes the iteration number, and $r$ basically controls the decaying speed of $s$ along with the increasing iterations. We set $s_{start}=0.2$, $r=50$ for the coarse stage and $s_{start}=0.05$, $r=50$ for the fine stage.

For all the experimental results on the DTU~\citep{jensen2014large} dataset, our method is trained on the training set with around 90\% images for each scene, following~\citep{wang2021neus} for the splitting scheme, and the 10\% images are used for evaluation of the novel view synthesis task. We notice that the CD performance is only slightly influenced compared to training on the full dataset. 
For experiments on the BlendedMVS~\citep{yao2020blendedmvs} dataset, we use all the images for training.

\subsection{Details for baseline methods}
We include the following baseline methods for comparison:
\textbf{IDR}~\citep{yariv2020multiview} reconstructs high-quality surfaces with implicit representation based on foreground object masks and the corresponding mask loss. 
\textbf{NeuS}~\citep{wang2021neus} is a state-of-the-art approach that develops a volume rendering method for surface reconstruction, where the mask supervision is optional. Reconstruction results for NeuS are implemented with its official code~\footnote{https://github.com/Totoro97/NeuS} and the pre-trained models, and the novel view rendering results are provided by the authors.
\textbf{NeRF}~\citep{mildenhall2020nerf} first proposes to use the neural radiance field for novel view synthesis. Though not specifically designed for surface reconstruction, we can extract a noise geometry from a trained NeRF model with a selected threshold. In this paper, the reconstruction evaluation results for NeRF are directly taken from~\citep{wang2021neus} for a fair comparison, while we also implement NeRF with nerf-pytorch~\footnote{https://github.com/yenchenlin/nerf-pytorch} for novel view synthesis. 
\textbf{DVGO}~\citep{sun2021direct} accelerates NeRF with a hybrid representation. We use the official code~\footnote{https://github.com/sunset1995/DirectVoxGO} and implement DVGO-v2~\citep{sun2022improved} for comparison, which is 2-3 times faster than the DVGO-v1. Similarly, we select a threshold to extract the geometry from the density voxel grid, as to be introduced below.
Results for these methods with the aid of foreground object masks are presented in Table 1 of the main text.

\begin{table}[t]
  \centering
  \small
  \caption{Quantitative evaluation on the DTU dataset for novel view synthesis. Our method outperforms the baselines on all the three metrics.}
    \setlength\tabcolsep{3.2pt}
    \begin{tabular}{c|cccc|cccc|cccc}
    \toprule
    \toprule
    Metric & \multicolumn{4}{c|}{PSNR$\uparrow$}     & \multicolumn{4}{c|}{SSIM$\uparrow$}     & \multicolumn{4}{c}{LPIPS$\downarrow$} \\
    \midrule
    Scan  & NeRF  & DVGO  & NeuS  & Ours  & NeRF  & DVGO  & NeuS  & Ours  & NeRF  & DVGO  & NeuS  & Ours \\
    \midrule
    24    & 26.97 & 27.77 & 26.13 & \textbf{27.89} & 0.772 & 0.830 & 0.764 & \textbf{0.857} & 0.331 & 0.277 & 0.348 & \textbf{0.239} \\
    37    & 25.99 & 25.96 & 24.08 & \textbf{26.90} & 0.811 & 0.833 & 0.798 & \textbf{0.870} & 0.206 & 0.184 & 0.222 & \textbf{0.160} \\
    40    & 27.68 & 27.75 & 26.73 & \textbf{28.81} & 0.786 & 0.791 & 0.747 & \textbf{0.841} & 0.304 & 0.303 & 0.352 & \textbf{0.274} \\
    55    & 29.39 & 30.42 & 28.06 & \textbf{31.02} & 0.917 & 0.939 & 0.887 & \textbf{0.950} & 0.143 & 0.116 & 0.177 & \textbf{0.108} \\
    63    & 33.07 & 34.35 & 28.69 & \textbf{34.38} & 0.936 & 0.953 & 0.937 & \textbf{0.957} & 0.128 & 0.095 & 0.129 & \textbf{0.083} \\
    65    & 30.87 & 31.18 & 31.41 & \textbf{31.48} & 0.954 & 0.956 & 0.958 & \textbf{0.960} & 0.114 & 0.103 & 0.112 & \textbf{0.094} \\
    69    & 27.90 & 29.52 & 28.96 & \textbf{30.13} & 0.844 & 0.921 & 0.909 & \textbf{0.928} & 0.308 & 0.190 & 0.223 & \textbf{0.181} \\
    83    & 33.49 & 36.94 & 31.56 & \textbf{37.43} & 0.948 & \textbf{0.969} & 0.950 & 0.968 & 0.125 & \textbf{0.084} & 0.120 & \textbf{0.084} \\
    97    & 27.43 & 27.67 & 25.51 & \textbf{28.35} & 0.900 & 0.914 & 0.901 & \textbf{0.923} & 0.200 & 0.168 & 0.192 & \textbf{0.155} \\
    105   & 31.68 & 32.85 & 29.18 & \textbf{32.94} & 0.910 & 0.928 & 0.896 & \textbf{0.932} & 0.186 & 0.154 & 0.218 & \textbf{0.148} \\
    106   & 30.73 & 33.75 & 32.60 & \textbf{34.17} & 0.879 & 0.933 & 0.914 & \textbf{0.947} & 0.244 & 0.167 & 0.201 & \textbf{0.138} \\
    110   & 29.61 & \textbf{33.10} & 30.83 & 32.70 & 0.872 & \textbf{0.941} & 0.917 & 0.937 & 0.241 & \textbf{0.153} & 0.200 & \textbf{0.153} \\
    114   & 29.37 & 30.18 & 29.32 & \textbf{30.97} & 0.901 & 0.914 & 0.897 & \textbf{0.926} & 0.193 & 0.174 & 0.216 & \textbf{0.159} \\
    118   & 33.44 & 36.11 & 35.91 & \textbf{37.24} & 0.915 & 0.957 & 0.948 & \textbf{0.964} & 0.199 & 0.123 & 0.156 & \textbf{0.110} \\
    122   & 33.41 & 36.99 & 35.49 & \textbf{37.97} & 0.935 & 0.967 & 0.957 & \textbf{0.972} & 0.142 & 0.088 & 0.114 & \textbf{0.076} \\
    \midrule
    mean  & 30.07 & 31.64 & 29.63 & \textbf{32.16} & 0.885 & 0.916 & 0.892 & \textbf{0.929} & 0.204 & 0.159 & 0.199 & \textbf{0.144} \\
    \bottomrule
    \bottomrule
    \end{tabular}%
  \label{tab:psnr_all}%
  \vspace{-15pt}
\end{table}%

\begin{table}[t]
  \centering
  \scriptsize
  \caption{Quantitative evaluation on DTU dataset (without mask).}
   \setlength\tabcolsep{3.2pt}
    \begin{tabular}{l|cccccccccccccccc}
    \toprule
    \toprule
    Scan & 24    & 37    & 40    & 55    & 63    & 65    & 69    & 83    & 97    & 105   & 106   & 110   & 114   & 118   & 122   & mean \\
    \midrule
    Colmap~\citep{schonberger2016pixelwise} & 0.81  & 2.05  & 0.73  & 1.22  & 1.79  & 1.58  & 1.02  & 3.05  & 1.40  & 2.05  & 1.00  & 1.32  & 0.49  & 0.78  & 1.17  & 1.36 \\
    NeRF~\citep{mildenhall2020nerf} & 1.90  & 1.60  & 1.85  & 0.58  & 2.28  & 1.27  & 1.47  & 1.67  & 2.05  & 1.07  & 0.88  & 2.53  & 1.06  & 1.15  & 0.96  & 1.49 \\
    UNISURF~\citep{oechsle2021unisurf} & 1.32  & 1.36  & 1.72  & 0.44  & 1.35  & 0.79  & 0.80  & 1.49  & 1.37  & 0.89  & 0.59  & 1.47  & 0.46  & 0.59  & 0.62  & 1.02 \\
    VolSDF~\cite{yariv2021volume} & 1.14  & 1.26  & 0.81  & 0.49  & 1.25  & 0.70  & 0.72  & 1.29  & 1.18  & \textbf{0.70}  & 0.66  & \textbf{1.08}  & 0.42  & 0.61  & 0.55  & 0.86 \\
    NeuS~\citep{wang2021neus} & 1.00  & 1.37  & 0.93  & 0.43  & \textbf{1.10}  & \textbf{0.65}  & \textbf{0.57}  & 1.48  & 1.09  & 0.83  & \textbf{0.52}  & 1.20  & \textbf{0.35}  & \textbf{0.49}  & 0.54  & 0.84 \\
    Ours  & \textbf{0.72}  & \textbf{0.75}  & \textbf{0.47}  & \textbf{0.39}  & 1.47  & 0.76  & 0.81  & \textbf{1.02}  & \textbf{1.04}  & 0.92  & \textbf{0.52}  & 1.13  & 0.40  & 0.53  & \textbf{0.53}  & \textbf{0.76} \\
    \bottomrule
    \bottomrule
    \end{tabular}%
    \vspace{-15pt}
  \label{tab:dtu_womask}%
\end{table}%

We also include several baselines that do not rely on foreground object masks: 
\textbf{Colmap}~\citep{schonberger2016pixelwise} is a widely-used classical Multi-view stereo method. 
\textbf{UNISURF}~\citep{oechsle2021unisurf} uses the occupancy field to represent the geometry and improves reconstruction quality by shrinking the sample region of volume rendering during training.
\textbf{VolSDF}~\citep{yariv2021volume} defines the volume density function as Laplace’s cumulative distribution function (CDF) applied to a SDF representation for surface reconstruction.
We also compare with \textbf{NeRF}~\citep{mildenhall2020nerf} and \textbf{NeuS}~\citep{wang2021neus} under the without-mask setting.
All the results above are directly adopted from the original papers. Their comparisons to our method under this setting are shown in Table~\ref{tab:dtu_womask}.

\begin{figure}[t]
\begin{minipage}[t]{\textwidth}
 \begin{minipage}[t]{0.49\textwidth}
  \centering
  \makeatletter\def\@captype{table}\makeatother
  \small
  \caption{Ablation over the level selection for hierarchical geometry feature. The performance first increases together with the level and then converges after level 2.0.} 
  \small
     \setlength\tabcolsep{3.2pt}
    \begin{tabular}{c|ccccccc}
    \toprule
    \toprule
    level & 0     & 0.5   & 1     & 1.5   & 2     & 2.5   & 3 \\
    \midrule
    CD    & 0.98    & 0.75  & 0.74  & 0.74  & \textbf{0.72}  & 0.73  & \textbf{0.72} \\
    \bottomrule
    \bottomrule
    \end{tabular}
    \vspace{-10pt}
  \label{tab:grad_level}
  \end{minipage}
  \hspace{20pt}
  \begin{minipage}[t]{0.49\textwidth}
  \centering
   \makeatletter\def\@captype{table}\makeatother
       \caption{Ablation over the geometry feature design. It indicates that a combination of both \textit{Gradient} and \textit{SDF} produces the best result.} 
    \small
    \begin{tabular}{l|cccc}
    \toprule
    \toprule
    \small
    CD (mean) & 0.79  & 0.76  & 0.74  & \textbf{0.72} \\
    \midrule
    Gradient  &       & \checkmark  &       & \checkmark \\
    SDF &       &       & \checkmark  & \checkmark \\
    
    \bottomrule
    \bottomrule
    \end{tabular}%
    \vspace{-10pt}
  \label{tab:grad_sdf}%
  \vspace{-10pt}
  \end{minipage}
\end{minipage}
\end{figure}

\begin{table}[t]
  \centering
  \small 
  \caption{Ablation over different smoothness priors. Our gradient smoothness loss is proved effective by this quantitative evaluation.
  }
    \begin{tabular}{l|cccc}
     \toprule
    \toprule
    \small
    CD (mean) & 1.18  & 0.79  & 0.74  & \textbf{0.72} \\
    \midrule
    SDF TV  &       & \checkmark  &       & \checkmark \\
    Gradient smoothness loss &       &       & \checkmark  & \checkmark \\
    \bottomrule
    \bottomrule
    \end{tabular}%
  \label{tab:smooth_tv}%
  \vspace{-15pt}
\end{table}%

\begin{table}[t]
  \centering
  \caption{The effect of voxel grid resolution on reconstruction performance and training time. All the cases below are trained with the same settings except for the voxel grid resolution. 
  }
    \begin{tabular}{c|cccc}
    \toprule
    \toprule
    Resolution & $128^3$ & $192^3$ & $256^3$ & $320^3$ \\
    \midrule
    CD (mean)   & 0.79  & 0.75 & 0.72 & 0.73 \\
    Train time   & 11 mines  & 12 mins & 15 mins & 17 mins \\
    \bottomrule
    \bottomrule
    \end{tabular}%
  \label{tab:resolution}%
  \vspace{-15pt}
\end{table}%

\begin{figure}[t]
	\centering
	\includegraphics[width=0.8\linewidth]{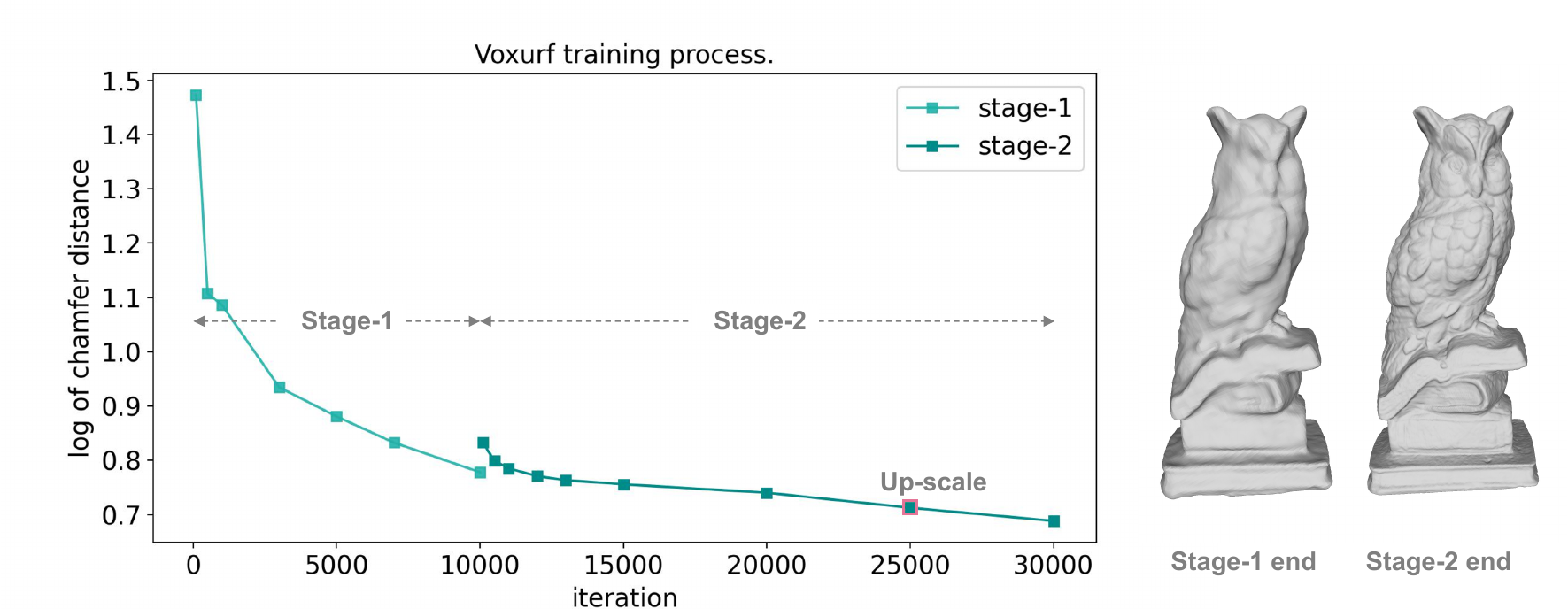}
	\vspace{-5pt}
	\caption{\small
	The two-stage training process of Voxurf. The number in the vertical axis is calculated by $\log_{10}(10 x)$ for better visualization.
	}
	\vspace{-15pt}
	\label{fig:two_stage}
\end{figure}

\begin{figure}[b]
	\centering
	\includegraphics[width=0.9\linewidth]{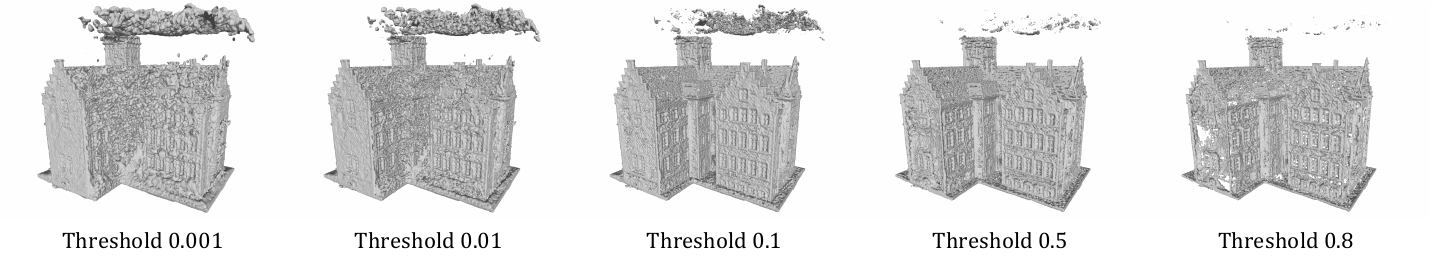}
	\vspace{-5pt}
	\caption{\small
	Comparisons of Alpha threshold selection for surface extraction from a trained DVGO~\citep{sun2021direct} model. A large threshold leads to holes and the incomplete surface, while a small one leads to floating noises above the surface.
	}
	\vspace{-5pt}
	\label{fig:threshold}
\end{figure}

\section{Additional experimental results}
\label{supp:more_exp}
\subsection{Novel view synthesis.}
\label{supp:psnr_all}
We report the results for novel view synthesis on the DTU dataset in Table~\ref{tab:psnr_all}. Our method outperforms the baselines in all three metrics, including PSNR, SSIM~\citep{wang2004image}, and LPIPS~\citep{zhang2018unreasonable} (VGG). 
Examples of rendered images at testing views are shown in Fig.~\ref{fig:render_1} and Fig.~\ref{fig:render_2} in Sec.~\ref{supp:more_vis}.

\subsection{Comparisons for the w/o mask setting.}
\label{supp:womask}
Our method can also work on cases where the background is not clean. Following NeRF++~\citep{zhang2020nerf++} and MipNeRF-360~\citep{barron2022mipnerf360}, we invert the background points outside the unit sphere into the unit sphere by $x' = x/r^2, y' = y/r^2, z' = z/r^2$, where $r=\sqrt{x^2 + y^2 + z^2}$.
We then represent the background with another density voxel grid, together with a feature grid and a shallow MLP. 
We report our results and comparisons to previous approaches in Table~\ref{tab:dtu_womask}.

\subsection{Additional ablation studies and analysis}
\noindent{\textbf{Ablation over the hierarchical geometry feature.}}
For hierarchical geometry feature design, we explore different design details, including the level selection and the effect of gradient and SDF value, as shown in Table~\ref{tab:grad_level} and Table~\ref{tab:grad_sdf}, respectively.

\noindent{\textbf{Ablation over smoothness priors.}}
\label{supp:smooth_tv}
(1) We introduce two regularization terms as smoothness priors during training, \ie the TV on the SDF voxel grid and a gradient smoothness prior. We carry out an ablation study on them during the fine training stage in Table~\ref{tab:smooth_tv}, where we reveal the effectiveness of our gradient smoothness loss via quantitative comparisons. \revision{(2)
Furthermore, we also evaluate the effectiveness of 3D convolution with a Gaussian kernel during coarse training and post-process. Experimental results show that the CD error increases from 0.72 to 0.74 if we remove Gaussian kernel during coarse training. And the it will be slightly reduced from 0.720 to 0.715 if we remove Gaussian kernel for post-process. The post-processing stage improves the visualization quality with a minor sacrifice of the quantitative performance.
}

\noindent{\textbf{Ablation over the voxel grid resolution.}}
\label{supp:resolution}
The voxel grid resolution denotes the number of voxels contained in $V^{(sdf)}$ of the fine training stage.
We study the effect of voxel grid resolution by keeping all the other settings to be the same, as shown in Table~\ref{tab:resolution}. 
Increasing the voxel grid resolution from $128^3$ to $192^3$ and from $192^3$ to $256^3$ consistently results in lower Chamfer Distance (CD) with longer training time.
However, the case with $320^3$ achieves a similar CD with $256^3$, and requires a higher training cost. Thus, we take the number of voxels to be $256^3$ as the default setting in the other experiments.

\noindent{\textbf{Two-stage training process.}}
\label{supp:two_stage}
Our method adopts a two-stage training pipeline. We show the curve of Chamfer Distance and the visualization result by the end of each stage in Fig.~\ref{fig:two_stage}. 
We show that 
1) we can obtain a coherent shape by the end of the Stage-1 (coarse training stage), while the performance is limited by the low resolution that the details are hard to be reconstructed;
2) the fine details are recovered by the end of Stage-2 (fine training stage) and the overall structure is consistent with the coarse shape of Stage-1.

\noindent{\textbf{Threshold selection.}}
\label{supp:threshold}
To extract the surface from a trained DVGO~\citep{sun2021direct} model, we can first obtain the alpha value for any point in the 3D space by density interpolation and activation. And then, we show how we select a proper alpha threshold when we extract the surface in Fig.~\ref{fig:threshold}. 
A small threshold like 0.001 and 0.01 usually results in noise areas floating above the surface, while a large one like 0.5 and 0.8 would lead to an incomplete surface with large holes. We thus select 0.1 as the alpha threshold that is adopted in this paper.

\section{DVGO+NeuS w/ Smoothness}
\label{supp:baseline_smooth}
\revision{Our method introduces several designs to boost the smoothness of the reconstruction results, including the 3D convolution with a Gaussian kernel $\mathcal{G}(V,k_g,\sigma_g)$ for coarse stage training (not adopted in fine stage) (Sec. 5.1), the gradient smoothness loss $\mathcal{L}_{smooth}(\nabla V^{(sdf)})$ (Sec. 5.3), a 3D convolution with a Gaussian kernel $\mathcal{G}(V,k_g,\sigma_g)$ for post-processing during inference (Sec. 5.3), and the SDF total variation loss $\mathcal{L}_{TV}(V^{(sdf)})$ (Sec. 5.3). }

\begin{figure}[t]
	\centering
	\includegraphics[width=0.9\linewidth]{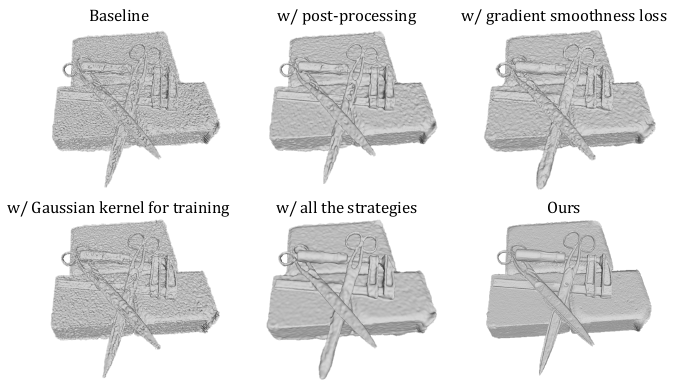}
	\caption{\small
	\revision{Qualitative comparisons of the DVGO+SDF baseline enhanced with different combinations of smoothness priors. A huge performance gap still exists when the strongest smoothness strategies are applied.}
	}
	\label{fig:baseline_w_smooth}
\end{figure}

\revision{Since we observe the combination of DVGO and NeuS (DVGO + NeuS) produces continuous but noisy surfaces, it is interesting to apply these smoothness designs to the baseline. In this way, we can more clearly verify the effectiveness of other technical contributions in Voxurf. Specifically, the SDF total variation loss (SDF TV) has already been adopted to DVGO + NeuS for better results in Fig. 1 and Table. 1. We further evaluate the Gaussian kernel for training, gradient smoothness loss, and Gaussian kernel for post-processing on DVGO + NeuS. 
The most simple approach to make surfaces smooth is adopting the Gaussian kernel for post-processing. 
As shown in Fig.~\ref{fig:baseline_w_smooth}, the surfaces clearly become smooth, but the chamfer distance (CD) becomes worse as in Table~\ref{tab:baseline_smooth}. As either the Gradient smooth loss or Gaussian kernel for training is leveraged, both the chamfer distance (CD) and surfaces become better. Notably, the Gaussian kernel for post-processing will not harm the performance here with these smooth priors (\ie, Gradient smoothness loss, Gaussian kernel for training) adopted. Finally, via combining all of these designs, we achieve smooth surfaces and lower CD. However, both the quantitative comparison and visualization show that Voxurf is significantly better than DVGO + NeuS w/ smoothness. Moreover, after adopting all these designs, the training time of DVGO + NeuS w/ smoothness increases from 12min to 15min which is slightly longer than Voxurf (14min), singe Voxurf does not need the Gaussian kernel for training in the second stage thanks to the other designs.}

\begin{table}[t]
  \centering
  \small 
  \caption{\small\revision{The experimental results of applying Gaussian kernel for post-processing, gradient smoothness loss, and Gaussian kernel for training on DVGO + NeuS. Notably, SDF TV has already been adopted as a default setting, thus it is not discussed here.}}
    \begin{tabular}{c|ccccccc}
     \toprule
    \toprule
    \small
    CD (mean) & 1.13  & 1.22  & 0.99  & 1.00 & 1.00 & 1.00 & 0.98 \\
    \midrule
    Post-processing  &       & \checkmark  &    & \checkmark &   & \checkmark & \checkmark\\
    Gradient smoothness loss &       &       & \checkmark  & \checkmark & & & \checkmark\\
    Gaussian kernel for training &       &       &   &   & \checkmark & \checkmark & \checkmark\\
    \midrule
    Voxurf & \multicolumn{7}{c}{\textbf{0.72}} \\
    \bottomrule
    \bottomrule
    \end{tabular}%
  \label{tab:baseline_smooth}%
\end{table}%

\section{Discussions on the assumption of color-geometry dependency}
\label{supp:color_geometry}
\begin{figure}[t]
	\centering
	\includegraphics[width=0.9\linewidth]{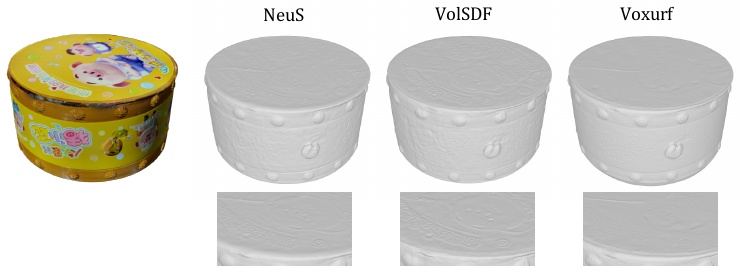}
	\caption{\small
	\revision{An example where the surface color is not correlated with geometry. The assumption of the color-geometry dependency leads to relief-like structures in VolSDF~\citep{yariv2021volume}, NeuS~\citep{wang2021neus}, and Voxurf, where our method is \textbf{least} affected by the side effect. It reveals that this problem can be mostly alleviated by multi-view consistency in an accurate reconstruction.
	}}
	\label{fig:side_effect}
\end{figure}

\revision{The assumption of color-geometry dependency is based on the idea of shape-from-shading, which has been proven effective in surface reconstruction by previous approaches~\citep{yariv2020multiview,yariv2021volume}. This technique generally does more good than harm, while side effects do exist in some cases where the surface texture is not correlated with the geometry/normal. We carry out several experiments with an example in Fig.~\ref{fig:side_effect}, where we can observe obvious relief-like structures on a plane surface caused by the texture.}

\revision{This is a common problem shared by most of the recent neural surface reconstruction methods~\citep{oechsle2021unisurf,yariv2020multiview,yariv2021volume,wang2021neus}.
Nevertheless, this problem can be largely alleviated via multi-view consistency when the geometry and color fields are well-trained with enough input views. In Fig.~\ref{fig:side_effect}, we observe that our method is better at overcoming the problem than previous state-of-the-art methods, being least affected by the textures. Fully addressing this side effect is out of the scope of this work and would be investigated in the future.}

\begin{figure}[t]
	\centering
	\includegraphics[width=0.99\linewidth]{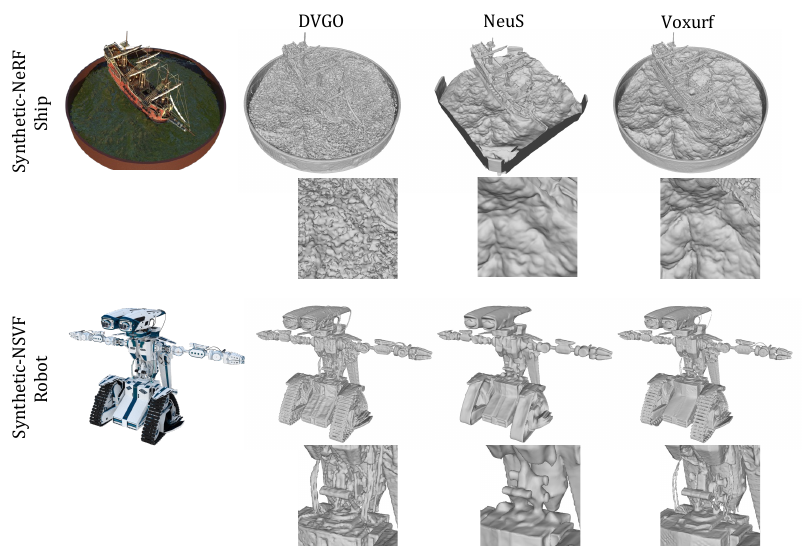}
	\vspace{-5pt}
	\caption{\small
	\revision{Qualitative comparisons on the \textit{Synthetic-NeRF} and \textit{Synthetic-NSVF} dataset.}
	}
	\vspace{-5pt}
	\label{fig:synthetic_and_nsvf}
\end{figure}

\section{Evaluations on new datasets}
\label{supp:more_datasets}

\revision{We further evaluate our method on other datasets, namely \textit{Synthetic-NeRF}~\citep{mildenhall2020nerf} and \textit{Synthetic-NSVF}~\citep{liu2020neural}, and a qualitative comparison can be found in Fig.~\ref{fig:synthetic_and_nsvf}. 
Compared with the baseline methods, our method shows superior performance, which does not suffer from the heavy noise as DVGO~\citep{sun2021direct} while producing far more accurate thin structures than NeuS~\citep{wang2021neus}.}

\revision{These datasets rarely appear in previous benchmarks of surface reconstruction, because they do NOT have publicly available 3D ground truth for evaluation. 
The average PSNRs of our method on \textit{Synthetic-NeRF} and \textit{Synthetic-NSVF} are 32.38 and 35.18, where DVGO achieves 31.95 and 35.08, respectively. Considering the time limitation, we will provide a full comparison of both geometry and novel view synthesis on multiple methods in the final version.
Results for more datasets (\eg, deepvoxels~\citep{sitzmann2019deepvoxels} and Tanks-and-Temples~\citep{Knapitsch2017tat}) will also be included in the final version.}

\section{Qualitative Results on Large Scenes}
\label{supp:large_scale}
\begin{figure}[t]
	\centering
	\includegraphics[width=0.9\linewidth]{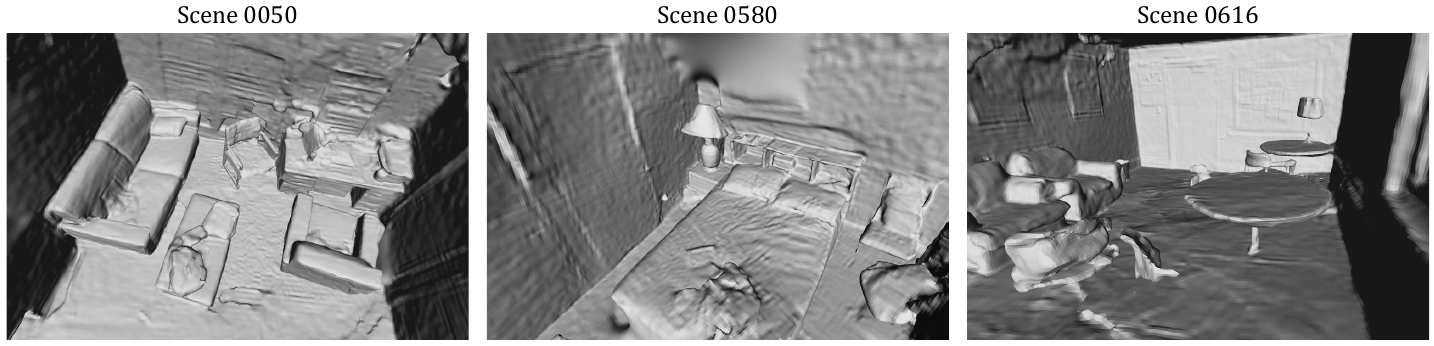}
	\vspace{-5pt}
	\caption{\small
	\revision{Qualitative results on large scenes, \eg, the ScanNet dataset.}
	}
	\vspace{-5pt}
	\label{fig:scannet}
\end{figure}

\revision{We further evaluate Voxurf on large scenes, \ie, Scannet dataset~\citep{dai2017scannet}. Following MonoSDF~\citep{Yu2022MonoSDF}, we adopt depth and normal estimated by MiDaS~\citep{Ranftl2022} as extra supervision. In Fig.~\ref{fig:scannet}, we show the qualitative results on scene 0050, 0580, and 0616 of Scannet. The results show that Voxurf can basically model the scene with an accurate layout and some details. However, it still performs unsatisfactory in reconstructing planes, \eg, floor and wall. We conjecture that the under-constrained voxel grids are not suitable to reconstruct from sparse-view observations like ScanNet videos. It is an import topic for future works.}

\section{Comparisons with more methods}
\label{supp:more_methods}
\revision{We further include Point-NeRF~\citep{xu2022point} for a comparison, which is an impressive work on NeRF acceleration based on an innovative neural point representation. This approach is proposed for the NVS task, and we manage to obtain the neural points after the finetuning stage and use Poisson surface reconstruction ~\citep{kazhdan2006poisson} to extract a surface from the points. 
The DTU performance on NVS task of Point-NeRF and our method are not directly comparable since they use different scenes and splits for training and testing. Therefore, we currently compare surface reconstruction results on the only one shared scene as shown in Fig.~\ref{fig:synthetic_and_nsvf}. It can be seen that the neural points after fine-tuning can roughly represent the surface of the scene, while the CD error (0.56) is obviously higher than ours (0.37). When we use Poisson surface reconstruction~\citep{kazhdan2006poisson}, a widely used method to mesh a point cloud, to recover a water-tight surface with the neural points, the CD error substantially increases to 1.77.
We will try to adopt Point-NeRF in the benchmark and perform a comprehensive evaluation on both surface reconstruction and rendering in the final version.}

\begin{figure}[t]
	\centering
	\includegraphics[width=0.9\linewidth]{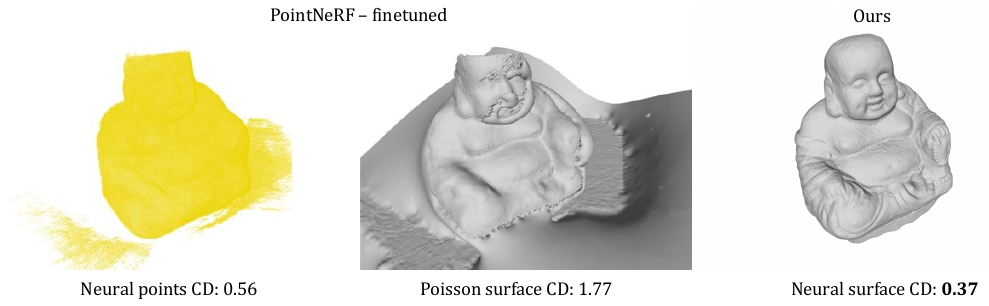}
	\vspace{-5pt}
	\caption{\small
	\revision{A comparison with Point-NeRF~\citep{xu2022point} in surface reconstruction. We use the Open3D Libarary~\citep{Zhou2018open3d} to perform Poisson surface reconstruction from the point cloud.}
	}
	\vspace{-5pt}
	\label{fig:synthetic_and_nsvf}
\end{figure}

\section{Limitations}
\label{supp:limitations}
\revision{Here we discuss the limitations and further research directions of our Voxurf.}

\revision{(1) Current neural surface reconstruction approaches~\citep{oechsle2021unisurf,yariv2020multiview,yariv2021volume,wang2021neus}, including our Voxurf, are built upon color-geometry dependence. Although the assumption does more good than harm, it sometimes causes side effects, \eg, a plane with printed textures will lead to relief-like structures. As discussed in Sec.~\ref{supp:color_geometry} and Fig.~\ref{fig:side_effect}, Voxurf is relatively less affected by such cases due to the accurate surface reconstruction enhanced by multi-view consistency, but the undesirable artifacts still exist. It is valuable to explore how to tackle this problem in the future.}

\revision{(2) Voxurf follows the NeRF-based techniques that represent 3D objects with view-conditioned emitted radiance.
This design is insufficient to reconstruct an accurate surface with less texture, strong reflection, and transparency due to strong ambiguity. This is another problem we can explore with Voxurf.}

\section{Additional qualitative comparisons}
\label{supp:more_vis}

\begin{figure}[h]
	\centering
	\includegraphics[width=1.0\linewidth]{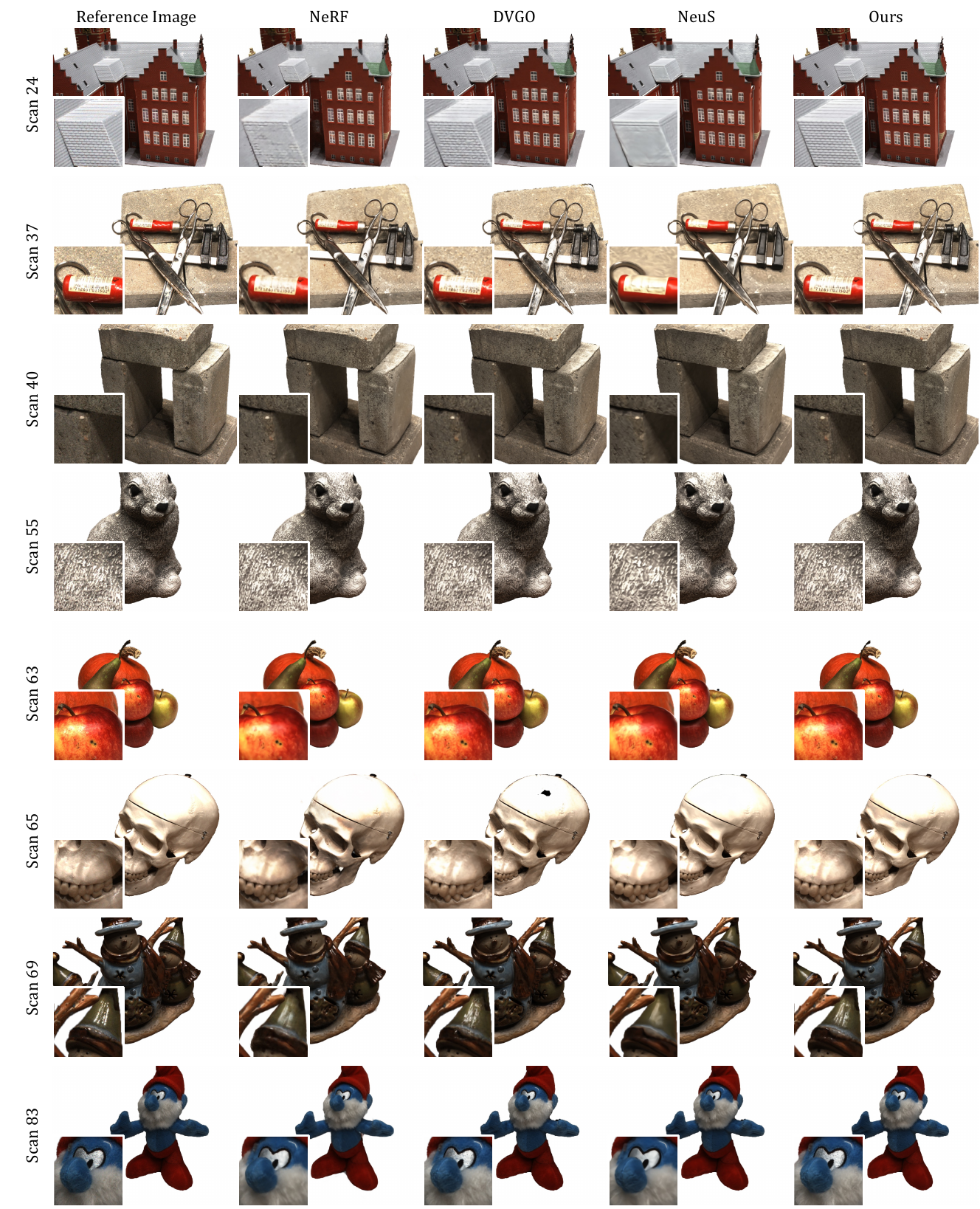}
	\vspace{-5pt}
	\caption{\small
	Qualitative comparisons on DTU for novel view synthesis. (Part 1/2)
	}
	\vspace{-5pt}
	\label{fig:render_1}
\end{figure}

\begin{figure}[t]
	\centering
	\includegraphics[width=1.0\linewidth]{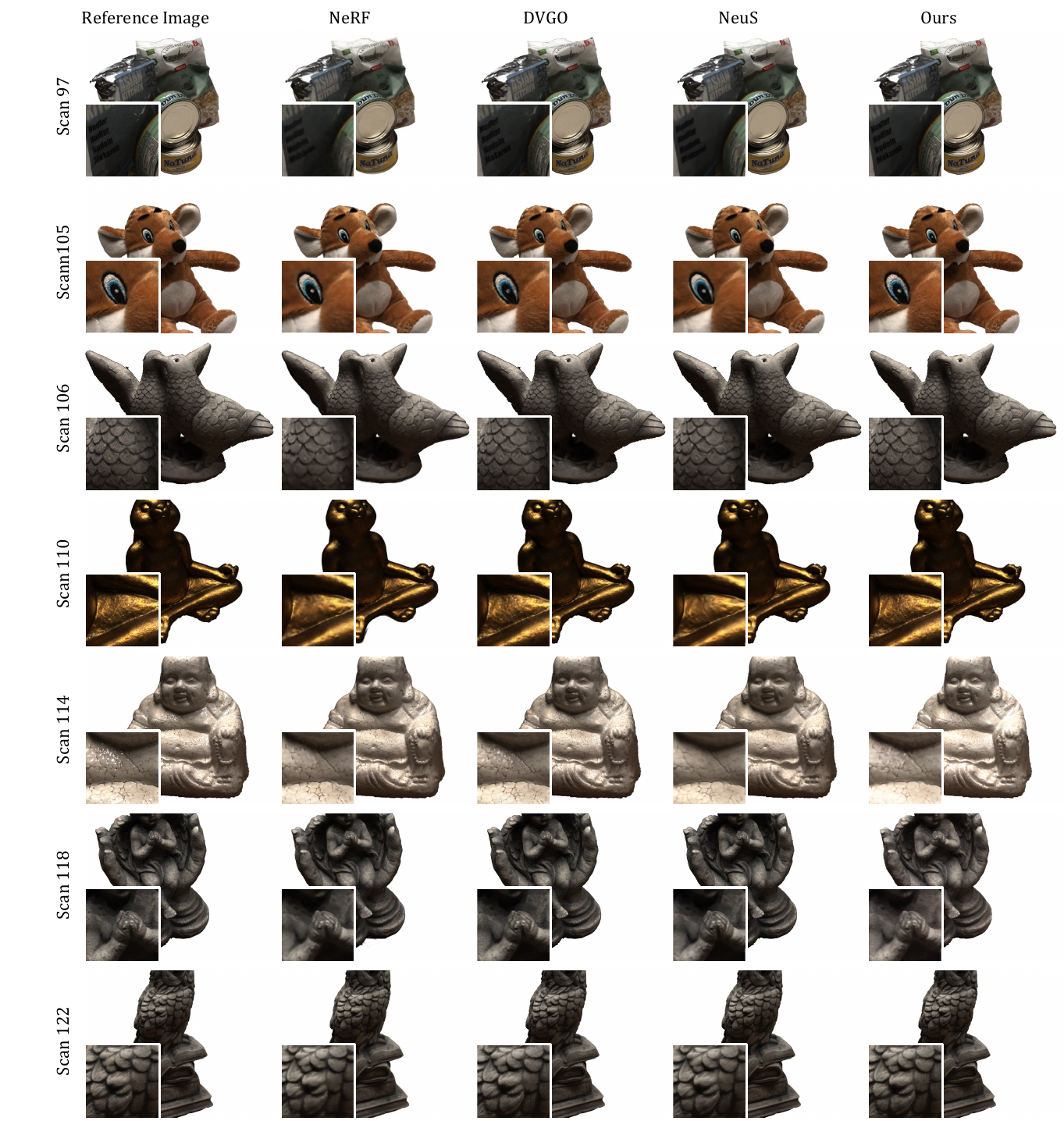}
	\vspace{-5pt}
	\caption{\small
	Qualitative comparisons on DTU for novel view synthesis. (Part 2/2)
	}
	\vspace{-5pt}
	\label{fig:render_2}
\end{figure}

\begin{figure}[t]
	\centering
	\includegraphics[width=1.0\linewidth]{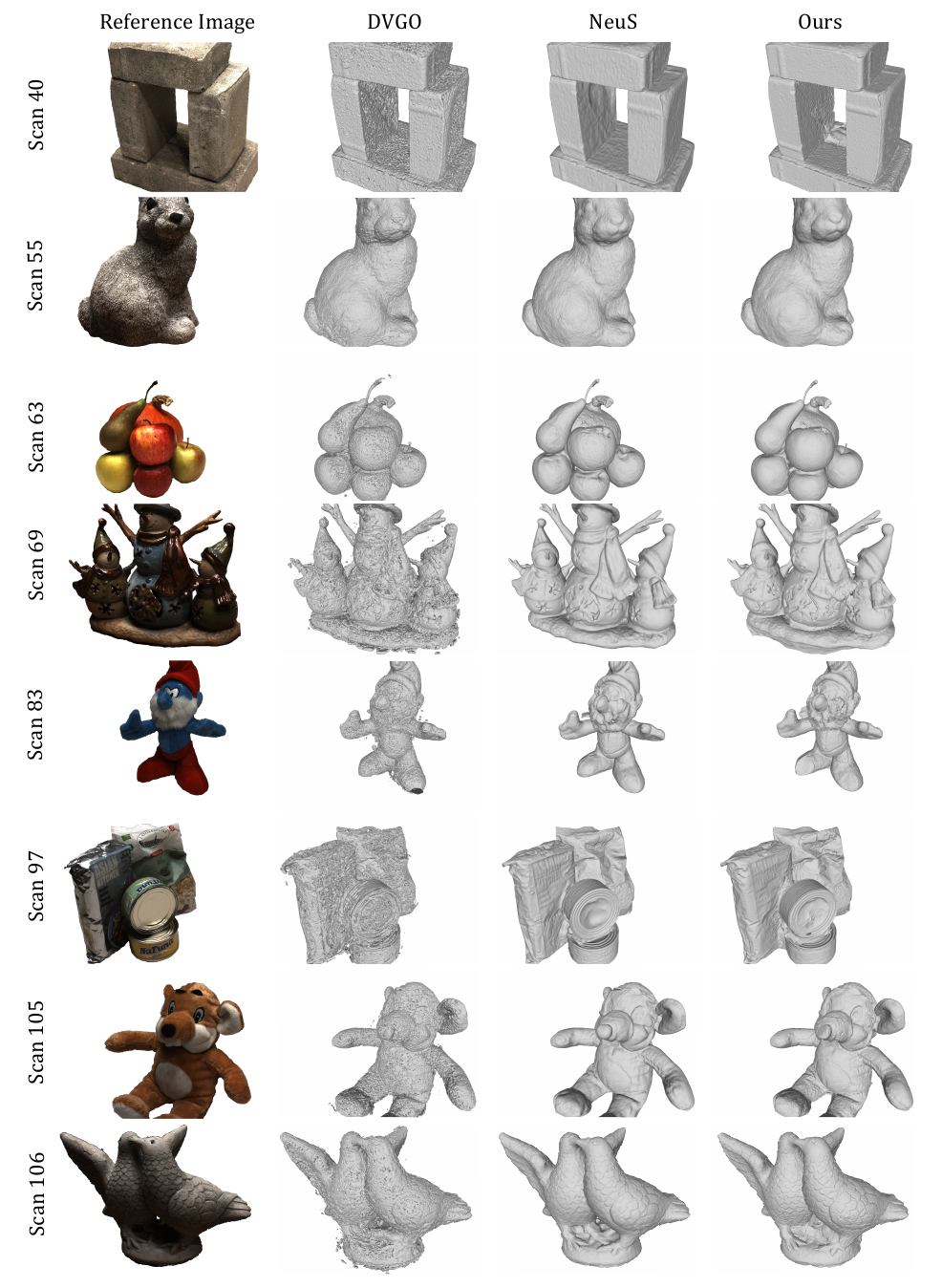}
	\vspace{-5pt}
	\caption{\small
	Additional surface reconstruction comparisons on DTU. (Part 1/2)
	}
	\vspace{-5pt}
	\label{fig:supp_dtu_1}
\end{figure}

\begin{figure}[t]
	\centering
	\includegraphics[width=1.0\linewidth]{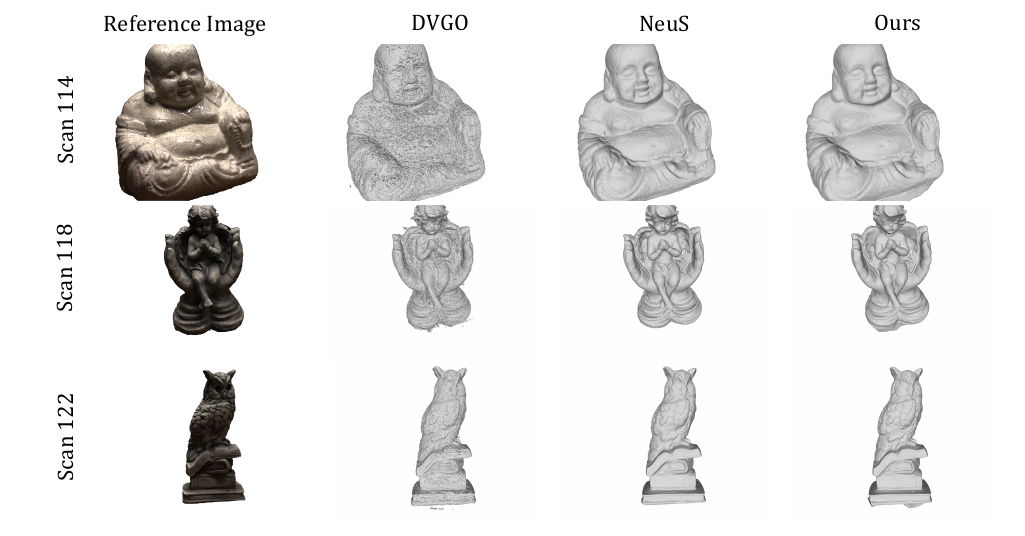}
	\vspace{-5pt}
	\caption{\small
	Additional surface reconstruction comparisons on DTU. (Part 2/2)
	}
	\vspace{-5pt}
	\label{fig:supp_dtu_2}
\end{figure}

\begin{figure}[t]
	\centering
	\includegraphics[width=1.0\linewidth]{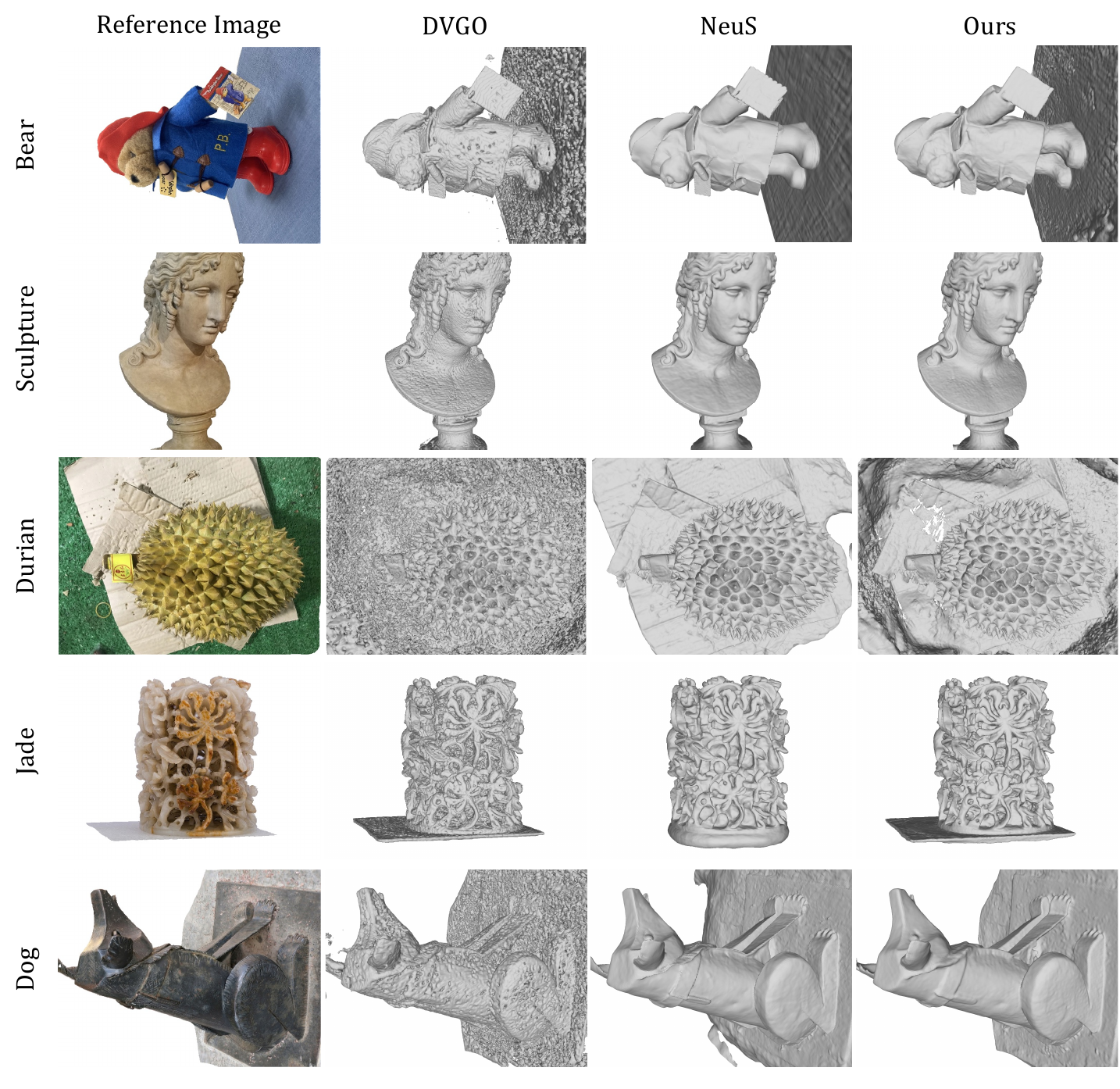}
	\vspace{-5pt}
	\caption{\small
	Additional surface reconstruction comparisons on BlendedMVS.
	}
	\vspace{-5pt}
	\label{fig:supp_bmvs}
\end{figure}

Finally, we show the qualitative comparisons for novel view synthesis in Fig.~\ref{fig:render_1} and Fig.~\ref{fig:render_2}, and we show additional surface reconstruction results in Fig~\ref{fig:supp_dtu_1}, Fig~\ref{fig:supp_dtu_2}, and Fig~\ref{fig:supp_bmvs}.

\clearpage
\bibliography{iclr2023_conference}

\begin{thebibliography}{48}
\providecommand{\natexlab}[1]{#1}
\providecommand{\url}[1]{\texttt{#1}}
\expandafter\ifx\csname urlstyle\endcsname\relax
  \providecommand{\doi}[1]{doi: #1}\else
  \providecommand{\doi}{doi: \begingroup \urlstyle{rm}\Url}\fi

\bibitem[Atzmon et~al.(2019)Atzmon, Haim, Yariv, Israelov, Maron, and
  Lipman]{atzmon2019controlling}
Matan Atzmon, Niv Haim, Lior Yariv, Ofer Israelov, Haggai Maron, and Yaron
  Lipman.
\newblock Controlling neural level sets.
\newblock \emph{Advances in Neural Information Processing Systems}, 32, 2019.

\bibitem[Barron et~al.(2022)Barron, Mildenhall, Verbin, Srinivasan, and
  Hedman]{barron2022mipnerf360}
Jonathan~T. Barron, Ben Mildenhall, Dor Verbin, Pratul~P. Srinivasan, and Peter
  Hedman.
\newblock Mip-nerf 360: Unbounded anti-aliased neural radiance fields.
\newblock \emph{CVPR}, 2022.

\bibitem[Chen et~al.(2022)Chen, Xu, Geiger, Yu, and Su]{chen2022tensorf}
Anpei Chen, Zexiang Xu, Andreas Geiger, Jingyi Yu, and Hao Su.
\newblock Tensorf: Tensorial radiance fields.
\newblock In \emph{European Conference on Computer Vision}, 2022.

\bibitem[Chen \& Zhang(2019)Chen and Zhang]{chen2019learning}
Zhiqin Chen and Hao Zhang.
\newblock Learning implicit fields for generative shape modeling.
\newblock In \emph{Proceedings of the IEEE/CVF Conference on Computer Vision
  and Pattern Recognition}, pp.\  5939--5948, 2019.

\bibitem[Dai et~al.(2017)Dai, Chang, Savva, Halber, Funkhouser, and
  Nie{\ss}ner]{dai2017scannet}
Angela Dai, Angel~X. Chang, Manolis Savva, Maciej Halber, Thomas Funkhouser,
  and Matthias Nie{\ss}ner.
\newblock Scannet: Richly-annotated 3d reconstructions of indoor scenes.
\newblock In \emph{Proc. Computer Vision and Pattern Recognition (CVPR), IEEE},
  2017.

\bibitem[Darmon et~al.(2022)Darmon, Bascle, Devaux, Monasse, and
  Aubry]{francois2021warping}
Fran{\c{c}}ois Darmon, B{\'{e}}n{\'{e}}dicte Bascle, Jean{-}Cl{\'{e}}ment
  Devaux, Pascal Monasse, and Mathieu Aubry.
\newblock Improving neural implicit surfaces geometry with patch warping.
\newblock In \emph{Proceedings of the IEEE/CVF International Conference on
  Computer Vision}, 2022.

\bibitem[Fang et~al.(2022)Fang, Yi, Wang, Xie, Zhang, Liu, Nie{\ss}ner, and
  Tian]{fang2022tineuvox}
Jiemin Fang, Taoran Yi, Xinggang Wang, Lingxi Xie, Xiaopeng Zhang, Wenyu Liu,
  Matthias Nie{\ss}ner, and Qi~Tian.
\newblock Fast dynamic radiance fields with time-aware neural voxels.
\newblock \emph{arxiv:2205.15285}, 2022.

\bibitem[Jensen et~al.(2014)Jensen, Dahl, Vogiatzis, Tola, and
  Aan{\ae}s]{jensen2014large}
Rasmus Jensen, Anders Dahl, George Vogiatzis, Engin Tola, and Henrik Aan{\ae}s.
\newblock Large scale multi-view stereopsis evaluation.
\newblock In \emph{Proceedings of the IEEE conference on computer vision and
  pattern recognition (CVPR)}, pp.\  406--413, 2014.

\bibitem[Jiang et~al.(2020)Jiang, Ji, Han, and Zwicker]{jiang2020sdfdiff}
Yue Jiang, Dantong Ji, Zhizhong Han, and Matthias Zwicker.
\newblock Sdfdiff: Differentiable rendering of signed distance fields for 3d
  shape optimization.
\newblock In \emph{The IEEE/CVF Conference on Computer Vision and Pattern
  Recognition (CVPR)}, June 2020.

\bibitem[Kazhdan et~al.(2006)Kazhdan, Bolitho, and Hoppe]{kazhdan2006poisson}
Michael Kazhdan, Matthew Bolitho, and Hugues Hoppe.
\newblock Poisson surface reconstruction.
\newblock In \emph{Proceedings of the fourth Eurographics symposium on Geometry
  processing}, volume~7, 2006.

\bibitem[Kellnhofer et~al.(2021)Kellnhofer, Jebe, Jones, Spicer, Pulli, and
  Wetzstein]{kellnhofer2021neural}
Petr Kellnhofer, Lars~C Jebe, Andrew Jones, Ryan Spicer, Kari Pulli, and Gordon
  Wetzstein.
\newblock Neural lumigraph rendering.
\newblock In \emph{Proceedings of the IEEE/CVF Conference on Computer Vision
  and Pattern Recognition}, pp.\  4287--4297, 2021.

\bibitem[Knapitsch et~al.(2017)Knapitsch, Park, Zhou, and
  Koltun]{Knapitsch2017tat}
Arno Knapitsch, Jaesik Park, Qian-Yi Zhou, and Vladlen Koltun.
\newblock Tanks and temples: Benchmarking large-scale scene reconstruction.
\newblock \emph{ACM Transactions on Graphics}, 36\penalty0 (4), 2017.

\bibitem[Liu et~al.(2020{\natexlab{a}})Liu, Gu, Lin, Chua, and
  Theobalt]{liu2020neural}
Lingjie Liu, Jiatao Gu, Kyaw~Zaw Lin, Tat-Seng Chua, and Christian Theobalt.
\newblock Neural sparse voxel fields.
\newblock \emph{Advances in Neural Information Processing Systems (NeurIPS)},
  2020{\natexlab{a}}.

\bibitem[Liu et~al.(2020{\natexlab{b}})Liu, Zhang, Peng, Shi, Pollefeys, and
  Cui]{liu2020dist}
Shaohui Liu, Yinda Zhang, Songyou Peng, Boxin Shi, Marc Pollefeys, and Zhaopeng
  Cui.
\newblock Dist: Rendering deep implicit signed distance function with
  differentiable sphere tracing.
\newblock In \emph{Proceedings of the IEEE/CVF Conference on Computer Vision
  and Pattern Recognition}, pp.\  2019--2028, 2020{\natexlab{b}}.

\bibitem[Lombardi et~al.(2019)Lombardi, Simon, Saragih, Schwartz, Lehrmann, and
  Sheikh]{lombardi2019neural}
Stephen Lombardi, Tomas Simon, Jason Saragih, Gabriel Schwartz, Andreas
  Lehrmann, and Yaser Sheikh.
\newblock Neural volumes: Learning dynamic renderable volumes from images.
\newblock \emph{ACM Transactions on Graphics (TOG)}, 2019.

\bibitem[Max(1995)]{max1995optical}
Nelson Max.
\newblock Optical models for direct volume rendering.
\newblock \emph{IEEE Transactions on Visualization and Computer Graphics},
  1\penalty0 (2):\penalty0 99--108, 1995.

\bibitem[Mescheder et~al.(2019)Mescheder, Oechsle, Niemeyer, Nowozin, and
  Geiger]{mescheder2019occupancy}
Lars Mescheder, Michael Oechsle, Michael Niemeyer, Sebastian Nowozin, and
  Andreas Geiger.
\newblock Occupancy networks: Learning 3d reconstruction in function space.
\newblock In \emph{Proceedings of the IEEE/CVF Conference on Computer Vision
  and Pattern Recognition}, pp.\  4460--4470, 2019.

\bibitem[Mildenhall et~al.(2019)Mildenhall, Srinivasan, Ortiz-Cayon, Kalantari,
  Ramamoorthi, Ng, and Kar]{mildenhall2019llff}
Ben Mildenhall, Pratul~P. Srinivasan, Rodrigo Ortiz-Cayon, Nima~Khademi
  Kalantari, Ravi Ramamoorthi, Ren Ng, and Abhishek Kar.
\newblock Local light field fusion: Practical view synthesis with prescriptive
  sampling guidelines.
\newblock \emph{ACM Transactions on Graphics (TOG)}, 2019.

\bibitem[Mildenhall et~al.(2020)Mildenhall, Srinivasan, Tancik, Barron,
  Ramamoorthi, and Ng]{mildenhall2020nerf}
Ben Mildenhall, Pratul~P Srinivasan, Matthew Tancik, Jonathan~T Barron, Ravi
  Ramamoorthi, and Ren Ng.
\newblock Nerf: Representing scenes as neural radiance fields for view
  synthesis.
\newblock In \emph{European conference on computer vision}, pp.\  405--421.
  Springer, 2020.

\bibitem[M\"uller et~al.(2022)M\"uller, Evans, Schied, and
  Keller]{mueller2022instant}
Thomas M\"uller, Alex Evans, Christoph Schied, and Alexander Keller.
\newblock Instant neural graphics primitives with a multiresolution hash
  encoding.
\newblock \emph{ACM Trans. Graph.}, 2022.

\bibitem[Niemeyer et~al.(2020)Niemeyer, Mescheder, Oechsle, and
  Geiger]{niemeyer2020differentiable}
Michael Niemeyer, Lars Mescheder, Michael Oechsle, and Andreas Geiger.
\newblock Differentiable volumetric rendering: Learning implicit 3d
  representations without 3d supervision.
\newblock In \emph{Proceedings of the IEEE/CVF Conference on Computer Vision
  and Pattern Recognition}, pp.\  3504--3515, 2020.

\bibitem[Oechsle et~al.(2021)Oechsle, Peng, and Geiger]{oechsle2021unisurf}
Michael Oechsle, Songyou Peng, and Andreas Geiger.
\newblock Unisurf: Unifying neural implicit surfaces and radiance fields for
  multi-view reconstruction.
\newblock In \emph{Proceedings of the IEEE/CVF International Conference on
  Computer Vision}, pp.\  5589--5599, 2021.

\bibitem[Park et~al.(2019)Park, Florence, Straub, Newcombe, and
  Lovegrove]{park2019deepsdf}
Jeong~Joon Park, Peter Florence, Julian Straub, Richard Newcombe, and Steven
  Lovegrove.
\newblock Deepsdf: Learning continuous signed distance functions for shape
  representation.
\newblock In \emph{Proceedings of the IEEE/CVF Conference on Computer Vision
  and Pattern Recognition}, pp.\  165--174, 2019.

\bibitem[Ranftl et~al.(2022)Ranftl, Lasinger, Hafner, Schindler, and
  Koltun]{Ranftl2022}
Ren\'{e} Ranftl, Katrin Lasinger, David Hafner, Konrad Schindler, and Vladlen
  Koltun.
\newblock Towards robust monocular depth estimation: Mixing datasets for
  zero-shot cross-dataset transfer.
\newblock \emph{IEEE Transactions on Pattern Analysis and Machine
  Intelligence}, 44\penalty0 (3), 2022.

\bibitem[Rudin \& Osher(1994)Rudin and Osher]{rudin1994total}
Leonid~I Rudin and Stanley Osher.
\newblock Total variation based image restoration with free local constraints.
\newblock In \emph{Proceedings of 1st international conference on image
  processing}, volume~1, pp.\  31--35. IEEE, 1994.

\bibitem[Saito et~al.(2019)Saito, Huang, Natsume, Morishima, Kanazawa, and
  Li]{saito2019pifu}
Shunsuke Saito, Zeng Huang, Ryota Natsume, Shigeo Morishima, Angjoo Kanazawa,
  and Hao Li.
\newblock Pifu: Pixel-aligned implicit function for high-resolution clothed
  human digitization.
\newblock In \emph{Proceedings of the IEEE/CVF International Conference on
  Computer Vision}, pp.\  2304--2314, 2019.

\bibitem[Sch{\"o}nberger et~al.(2016)Sch{\"o}nberger, Zheng, Frahm, and
  Pollefeys]{schonberger2016pixelwise}
Johannes~L Sch{\"o}nberger, Enliang Zheng, Jan-Michael Frahm, and Marc
  Pollefeys.
\newblock Pixelwise view selection for unstructured multi-view stereo.
\newblock In \emph{European Conference on Computer Vision}, pp.\  501--518.
  Springer, 2016.

\bibitem[Sitzmann et~al.(2019{\natexlab{a}})Sitzmann, Thies, Heide,
  Nie{\ss}ner, Wetzstein, and Zollhofer]{sitzmann2019deepvoxels}
Vincent Sitzmann, Justus Thies, Felix Heide, Matthias Nie{\ss}ner, Gordon
  Wetzstein, and Michael Zollhofer.
\newblock Deepvoxels: Learning persistent 3d feature embeddings.
\newblock In \emph{Proceedings of the IEEE/CVF Conference on Computer Vision
  and Pattern Recognition}, pp.\  2437--2446, 2019{\natexlab{a}}.

\bibitem[Sitzmann et~al.(2019{\natexlab{b}})Sitzmann, Zollh{\"o}fer, and
  Wetzstein]{sitzmann2019srns}
Vincent Sitzmann, Michael Zollh{\"o}fer, and Gordon Wetzstein.
\newblock Scene representation networks: Continuous 3d-structure-aware neural
  scene representations.
\newblock In \emph{Advances in Neural Information Processing Systems},
  2019{\natexlab{b}}.

\bibitem[Sun et~al.(2022{\natexlab{a}})Sun, Sun, and Chen]{sun2021direct}
Cheng Sun, Min Sun, and Hwann-Tzong Chen.
\newblock Direct voxel grid optimization: Super-fast convergence for radiance
  fields reconstruction.
\newblock In \emph{Proceedings of the IEEE/CVF International Conference on
  Computer Vision}, 2022{\natexlab{a}}.

\bibitem[Sun et~al.(2022{\natexlab{b}})Sun, Sun, and Chen]{sun2022improved}
Cheng Sun, Min Sun, and Hwann-Tzong Chen.
\newblock Improved direct voxel grid optimization for radiance fields
  reconstruction, 2022{\natexlab{b}}.

\bibitem[Tancik et~al.(2020)Tancik, Srinivasan, Mildenhall, Fridovich-Keil,
  Raghavan, Singhal, Ramamoorthi, Barron, and Ng]{tancik2020fourier}
Matthew Tancik, Pratul Srinivasan, Ben Mildenhall, Sara Fridovich-Keil, Nithin
  Raghavan, Utkarsh Singhal, Ravi Ramamoorthi, Jonathan Barron, and Ren Ng.
\newblock Fourier features let networks learn high frequency functions in low
  dimensional domains.
\newblock \emph{Advances in Neural Information Processing Systems},
  33:\penalty0 7537--7547, 2020.

\bibitem[Toussaint et~al.(2022)Toussaint, Genisson, and
  Franco]{toussaint2022hal}
Briac Toussaint, Maxime Genisson, and Jean-S{\'e}bastien Franco.
\newblock {Fast Gradient Descent for Surface Capture Via Differentiable
  Rendering}.
\newblock In \emph{{3DV 2022 - International Conference on 3D Vision}}, pp.\
  1--10, September 2022.

\bibitem[Wang et~al.(2022)Wang, Bleja, and Agapito]{wang2022gosurf}
Jingwen Wang, Tymoteusz Bleja, and Lourdes Agapito.
\newblock Go-surf: Neural feature grid optimization for fast, high-fidelity
  rgb-d surface reconstruction.
\newblock In \emph{2022 International Conference on 3D Vision (3DV)}. IEEE,
  2022.

\bibitem[Wang et~al.(2021)Wang, Liu, Liu, Theobalt, Komura, and
  Wang]{wang2021neus}
Peng Wang, Lingjie Liu, Yuan Liu, Christian Theobalt, Taku Komura, and Wenping
  Wang.
\newblock Neus: Learning neural implicit surfaces by volume rendering for
  multi-view reconstruction.
\newblock \emph{Advances in Neural Information Processing Systems (NeurIPS)},
  2021.

\bibitem[Wang et~al.(2004)Wang, Bovik, Sheikh, and Simoncelli]{wang2004image}
Zhou Wang, Alan~C Bovik, Hamid~R Sheikh, and Eero~P Simoncelli.
\newblock Image quality assessment: from error visibility to structural
  similarity.
\newblock \emph{IEEE transactions on image processing}, 13\penalty0
  (4):\penalty0 600--612, 2004.

\bibitem[Wizadwongsa et~al.(2021)Wizadwongsa, Phongthawee, Yenphraphai, and
  Suwajanakorn]{wizadwongsa2021nex}
Suttisak Wizadwongsa, Pakkapon Phongthawee, Jiraphon Yenphraphai, and Supasorn
  Suwajanakorn.
\newblock Nex: Real-time view synthesis with neural basis expansion.
\newblock In \emph{Proceedings of the IEEE/CVF Conference on Computer Vision
  and Pattern Recognition}, pp.\  8534--8543, 2021.

\bibitem[Xu et~al.(2022)Xu, Xu, Philip, Bi, Shu, Sunkavalli, and
  Neumann]{xu2022point}
Qiangeng Xu, Zexiang Xu, Julien Philip, Sai Bi, Zhixin Shu, Kalyan Sunkavalli,
  and Ulrich Neumann.
\newblock Point-nerf: Point-based neural radiance fields.
\newblock In \emph{Proceedings of the IEEE/CVF Conference on Computer Vision
  and Pattern Recognitio}, 2022.

\bibitem[Yao et~al.(2020)Yao, Luo, Li, Zhang, Ren, Zhou, Fang, and
  Quan]{yao2020blendedmvs}
Yao Yao, Zixin Luo, Shiwei Li, Jingyang Zhang, Yufan Ren, Lei Zhou, Tian Fang,
  and Long Quan.
\newblock Blendedmvs: A large-scale dataset for generalized multi-view stereo
  networks.
\newblock In \emph{Proceedings of the IEEE/CVF Conference on Computer Vision
  and Pattern Recognition}, pp.\  1790--1799, 2020.

\bibitem[Yariv et~al.(2020)Yariv, Kasten, Moran, Galun, Atzmon, Ronen, and
  Lipman]{yariv2020multiview}
Lior Yariv, Yoni Kasten, Dror Moran, Meirav Galun, Matan Atzmon, Basri Ronen,
  and Yaron Lipman.
\newblock Multiview neural surface reconstruction by disentangling geometry and
  appearance.
\newblock \emph{Advances in Neural Information Processing Systems (NeurIPS)},
  33, 2020.

\bibitem[Yariv et~al.(2021)Yariv, Gu, Kasten, and Lipman]{yariv2021volume}
Lior Yariv, Jiatao Gu, Yoni Kasten, and Yaron Lipman.
\newblock Volume rendering of neural implicit surfaces.
\newblock In \emph{Thirty-Fifth Conference on Neural Information Processing
  Systems}, 2021.

\bibitem[Yu et~al.(2022{\natexlab{a}})Yu, Fridovich-Keil, Tancik, Chen, Recht,
  and Kanazawa]{yu2021plenoxels}
Alex Yu, Sara Fridovich-Keil, Matthew Tancik, Qinhong Chen, Benjamin Recht, and
  Angjoo Kanazawa.
\newblock Plenoxels: Radiance fields without neural networks.
\newblock In \emph{Proceedings of the IEEE/CVF International Conference on
  Computer Vision}, 2022{\natexlab{a}}.

\bibitem[Yu et~al.(2022{\natexlab{b}})Yu, Peng, Niemeyer, Sattler, and
  Geiger]{Yu2022MonoSDF}
Zehao Yu, Songyou Peng, Michael Niemeyer, Torsten Sattler, and Andreas Geiger.
\newblock Monosdf: Exploring monocular geometric cues for neural implicit
  surface reconstruction.
\newblock \emph{Advances in Neural Information Processing Systems (NeurIPS)},
  2022{\natexlab{b}}.

\bibitem[Zhang et~al.(2021)Zhang, Yao, and Quan]{zhang2021learning}
Jingyang Zhang, Yao Yao, and Long Quan.
\newblock Learning signed distance field for multi-view surface reconstruction.
\newblock \emph{International Conference on Computer Vision (ICCV)}, 2021.

\bibitem[Zhang et~al.(2020)Zhang, Riegler, Snavely, and
  Koltun]{zhang2020nerf++}
Kai Zhang, Gernot Riegler, Noah Snavely, and Vladlen Koltun.
\newblock Nerf++: Analyzing and improving neural radiance fields.
\newblock \emph{arXiv preprint arXiv:2010.07492}, 2020.

\bibitem[Zhang et~al.(2018)Zhang, Isola, Efros, Shechtman, and
  Wang]{zhang2018unreasonable}
Richard Zhang, Phillip Isola, Alexei~A Efros, Eli Shechtman, and Oliver Wang.
\newblock The unreasonable effectiveness of deep features as a perceptual
  metric.
\newblock In \emph{Proceedings of the IEEE conference on computer vision and
  pattern recognition}, pp.\  586--595, 2018.

\bibitem[Zhang et~al.(2022)Zhang, Bi, Sunkavalli, Su, and
  Xu]{zhang2022nerfusion}
Xiaoshuai Zhang, Sai Bi, Kalyan Sunkavalli, Hao Su, and Zexiang Xu.
\newblock Nerfusion: Fusing radiance fields for large-scale scene
  reconstruction.
\newblock In \emph{Proceedings of the IEEE/CVF Conference on Computer Vision
  and Pattern Recognition}, 2022.

\bibitem[Zhou et~al.(2018)Zhou, Park, and Koltun]{Zhou2018open3d}
Qian-Yi Zhou, Jaesik Park, and Vladlen Koltun.
\newblock {Open3D}: {A} modern library for {3D} data processing.
\newblock \emph{arXiv:1801.09847}, 2018.

\end{thebibliography}
\bibliographystyle{iclr2023_conference}

\end{document}